\documentclass[preprint,12pt,authoryear]{elsarticle}

\usepackage{helvet}
\usepackage{courier}

\usepackage{bm}
\usepackage{latexsym}
\usepackage{epsfig}
\usepackage{amsbsy}
\usepackage{array}
\usepackage{amssymb}
\usepackage{setspace}
\usepackage{multicol}
\usepackage{graphicx}
\usepackage{pgfplots}
\usepackage{amsmath,graphicx}
\usepackage{amssymb}
\usepackage{algorithm}
\usepackage{algorithmic}
\usepackage{subfigure}

\journal{Signal Processing}

\begin{document}

\begin{frontmatter}



\title{Class Specific or Shared? A Cascaded Dictionary Learning Framework for Image Classification}


\author[UPC]{Yan-Jiang Wang}
\ead{yjwang@upc.edu.cn}
\author[UPC]{Shuai Shao}
\ead{shuaishao@s.upc.edu.cn}
\author[UPC]{Rui Xu}
\ead{xddxxr@126.com}
\author[UPC]{Weifeng Liu}
\ead{liuwf@upc.edu.cn}
\author[UPC]{Bao-Di Liu\corref{cor1}}
\cortext[cor1]{PaperID: SIGPRO-S-19-02500 Corresponding author.}
\ead{thu.liubaodi@gmail.com}

\address[UPC]{College of Control Science and Engineering, China University of Petroleum (Huadong), Qingdao 266580, China}

\begin{abstract}
Dictionary learning methods can be split into: i) class specific dictionary learning ii) class shared dictionary learning. 
The difference between the two categories is how to use discriminative information. With the first category, samples of different classes are mapped into different subspaces, which leads to some redundancy with the class specific base vectors. 
While for the second category, the samples in each specific class can not be described accurately. In this paper, we first propose a novel class shared dictionary learning method named label embedded dictionary learning (LEDL). It is the improvement based on LCKSVD, which is easier to find out the optimal solution. Then we propose a novel framework named cascaded dictionary learning framework (CDLF) to combine the specific dictionary learning with shared dictionary learning to describe the feature to boost the performance of classification sufficiently. Extensive experimental results on six benchmark datasets illustrate that our methods are capable of achieving superior performance compared to several state-of-art classification algorithms.
\end{abstract}

\end{frontmatter}


\section{Introduction}
\label{Introduction}

In recent years, image classification has been a classical issue in pattern recognition. With advancement in theory, many image classification methods have been proposed~\cite{wright2009robust,liu2019group,zhang2011sparse,yang2009linear,liu2016face,liu2017class,aharon2006k,zhang2010discriminative,jiang2013label,chan2015pcanet,he2016deep}. In these methods, there is one category that contributes a lot for image classification which is the dictionary learning (DL) based method. DL is a generative model of which the concept was firstly proposed by Mallat \emph{et al.}~\cite{mallat1993matching}. A few years later, Olshausen \emph{et al.}~\cite{olshausen1996emergence,olshausen1997sparse} proposed an application of DL on natural images and then it has been widely used in many fields such as image denoising~\cite{li2018joint,li2012efficient}, image superresolution~\cite{gao2018self,jiang2019edge} and image classification~\cite{li2019discriminative,lin2018robust}. According to different ways of utilizing the discriminative information, DL methods can be split into two categories: i) class specific dictionary learning ii) class shared dictionary learning.

Class specific dictionary learning method utilizes the discriminative information by adding discrimination ability into a dictionary. The learned dictionary is for each class. This category can gain the representative feature information of a class. The feature information that most of the class's samples have is focused on, while only a few samples of the class have is ignored to some extent. That is to say, the learned dictionary has higher weight on the feature information which samples close to the distribution center, and lower weight on the feature information that samples off the center. With this method, some abnormal sample points are ignored so that the learned dictionary's robustness can be improved. There are many classical class specific dictionary learning algorithms that have been reported in recent years. For example, 
fisher discrimination dictionary learning (FDDL)~\cite{yang2011fisher} uses the fisher discrimination criterion to learn a structured dictionary for pattern classification. In this method, each sub-dictionary can well reconstruct the signals from the same class. Different from FDDL, projective dictionary pair learning (DPL)~\cite{gu2014projective} and sevral following works such as robust adaptive dictionary pair learning (RA-DPL)~\cite{sun2020discriminative}, twin-projective latent dictionary pairs learning (TP-DPL)~\cite{zhang2019learning} extend the regular dictionary learning to dictionary pair learning. This kind of approach achieves the goal of signal representation and discrimination by jointly learning a synthesis dictionary and an analysis dictionary. Moreover, discriminative Bayesian dictionary learning (DBDL)~\cite{akhtar2016discriminative} and joint embedding and dictionary learning framework (JEDL)~\cite{zhang2016sparse} are two effective methods based class specific DL for classification. Specifically, DBDL complete classification task by approaching a non-parametric Bayesian perspective and JEDL do it according to deliver a linear sparse codes auto-extractor and a multi-class classifier by simultaneously minimizing the sparse reconstruction, discriminative sparse-code, code approximation, and classification errors.

Due that the learned dictionary is for each class, the training samples of each class are mapped into a separate subspace. First, different dictionaries can be regarded as different subspaces, which ensure that the salient characteristics of each category are elegantly described ($\ell_1$ norm constraint would generate the salient characteristics), and the discriminative information among different categories is obtained obviously (for each category, most of the learned atoms have their category characteristics and only a few will overlap with other categories' atoms). Second, each atom of dictionaries is $\ell_2$ normalized to guarantee that all the atoms are distributed on a hyper-surface. And thus, the projection of the image features is under the same scale. Third, although different dictionaries can be regarded as different subspaces, these dictionaries are not unrelated. For example, for the face dataset, different categories usually have the same characteristics (e.g., similar facial structure). When learning the dictionaries, the atoms would have some intersections or overlaps, which ensure that different categories of projections are comparable.

\begin{figure}[t]
	\begin{center}
		\includegraphics[width=0.8\linewidth]{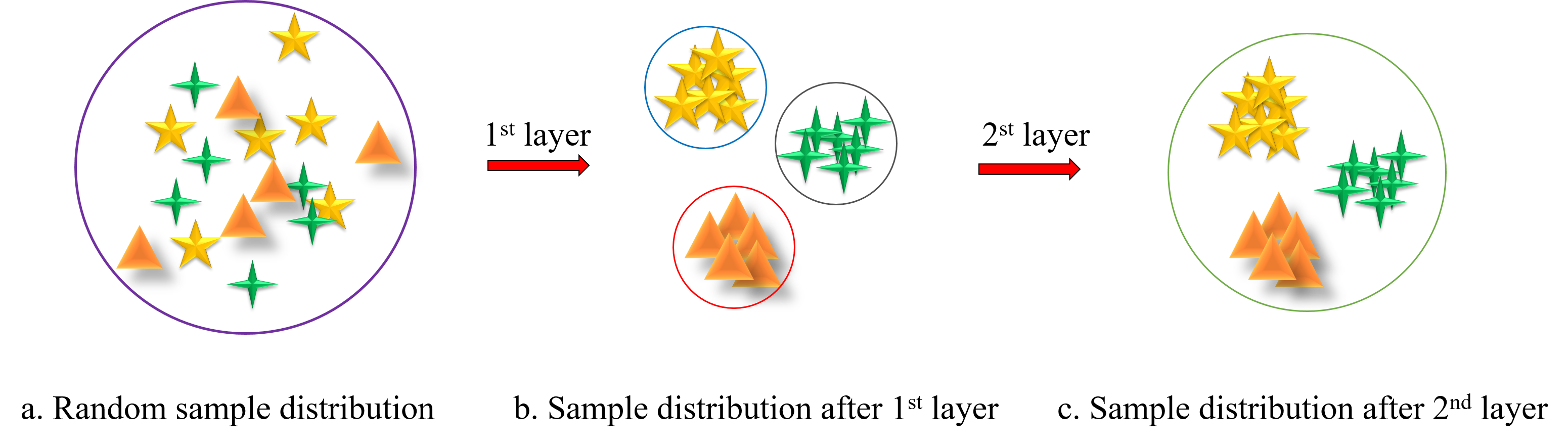}
	\end{center}
	\caption{Illustration of the variation of sample distribution. Different circles represent different subspaces.}
	\label{fig:Different_features}
\end{figure}

For class shared dictionary learning method, the discriminative information is directly embedded into the objective function to learn a dictionary for all classes. With this method, the training samples from all classes are mapped into one subspace. Hence, the representative feature information of all classes can be adopted. In recent years, many class shared dictionary learning methods have been proposed. For instance, label consistent K-SVD (LC-KSVD)~\cite{jiang2013label} proposed a new method to use a discriminative dictionary for sparse coding. In robust flexible discriminative dictionary learning (RFDDL)~\cite{zhang2019joint}, the labels are utilized flexibly to enhance the robust property to sparse errors and encoding the locality, reconstruction error and label consistency more accurately. And the joint robust factorization and projective dictionary learning (J-RFDL)~\cite{ren2020learning} discovered the hybrid salient low-rank and sparse representation in a factorized compressed space and use it improve the data representations. All of the above-mentioned methods which are based on class shared dictionary learning approaches to update the dictionary, have got an outstanding performance on classification tasks. However, it can not describe the samples in each specific class accurately so that the detailed features between samples in one class can not be found easily. 

In comparison to class specific dictionary learning and class shared dictionary learning, it is clear that the two methods have complementary advantages. It would help to obtain a significant boost in classification accuracy if the advantages of the two dictionary learning methods can be appropriately combined. In this paper, we first propose a novel class shared dictionary learning algorithm named label embedded dictionary learning (LEDL). This method introduces the $\ell_1$-norm regularization term to replace the $\ell_0$-norm regularization of LC-KSVD. In LC-KSVD, $\ell_0$-norm based sparse regularization term leads to the NP-hard problem. Even though many classical greedy methods such as orthogonal matching pursuit (OMP)~\cite{tropp2007signal} have solved this problem to some extent, it is usually to find the suboptimum sparse solution instead of the optimal sparse solution. In addition, the greedy method solves the global optimal problems by finding basis vectors in order of reconstruction errors from small to large until $T$ (the sparsity constraint factor) times. Thus, the initialized values are crucial. To this end, there are still many restrictions by using $\ell_0$-norm based sparse constraint. While our proposed LEDL is helpful to solve this problem. Moreover, compared with some dictionary learning methods without $\ell_0/\ell_0$-norm regularization term, such as robust label embedding projective dictionary learning (LE-PDL)~\cite{jiang2017robust} and scalable locality-constrained projective dictionary learning (LC-PDL)~\cite{zhang2019scalable}, our proposed methods are more general. Despite the fact that these methods have achieved good performance, they may lead to some singular values which come from the non-full rank matrices in some datasets. After that, we propose a novel framework named cascaded dictionary learning framework (CDLF) to combine a class specific dictionary learning method with a class shared dictionary learning method together. 

\begin{figure}[t]
	\begin{center}
		\includegraphics[width=0.8\linewidth]{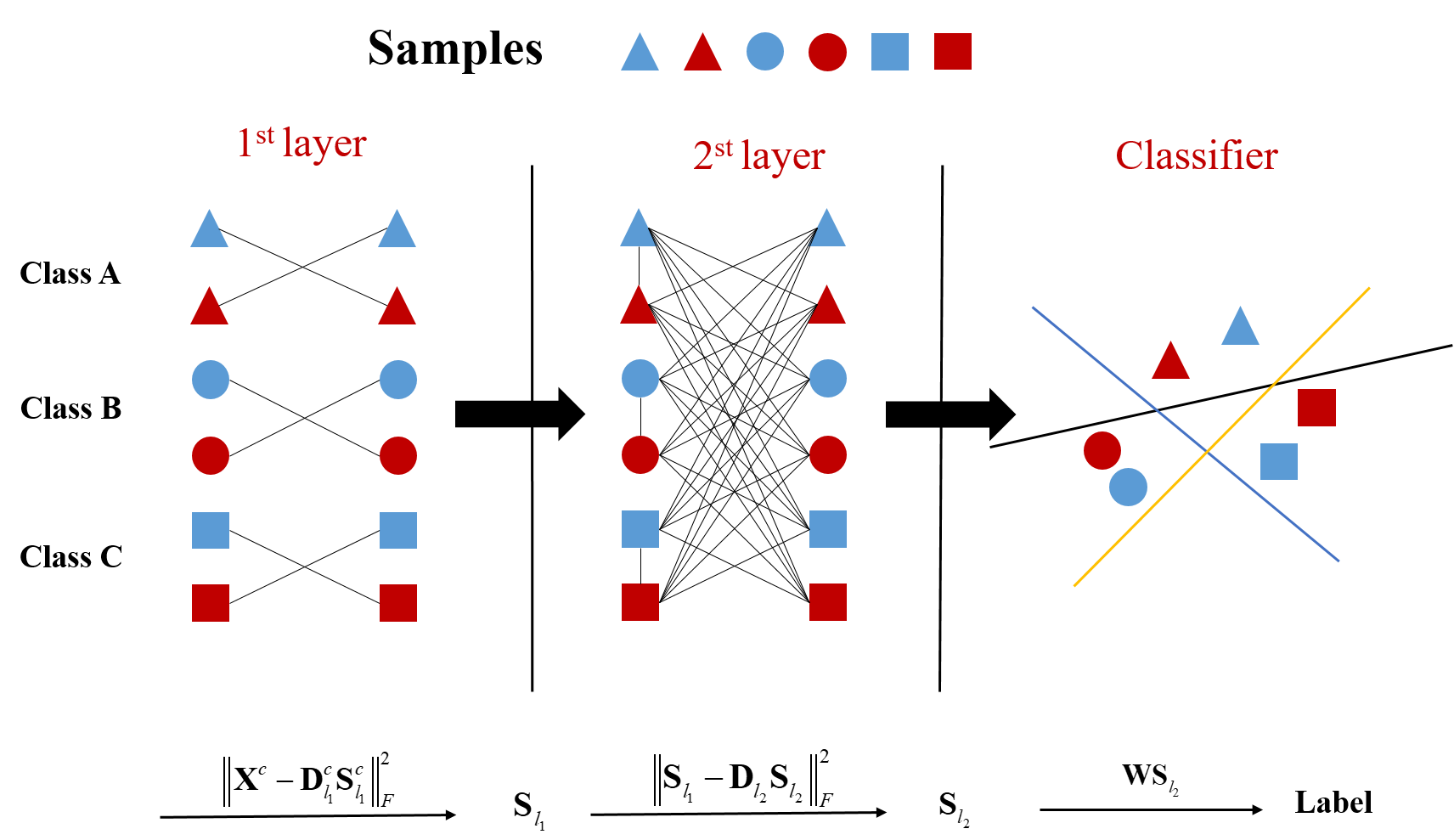}
	\end{center}
	\caption{
		Illustration of the cascaded dictionary learning framework (CDLF). In the first layer, accurate subspaces are created for each class by class specific dictionary method. Therefore the samples from the same class are clustered together. 
		In the second layer, the label embeded dictionary learning method establishes a discriminative subspace for all classes. $\bf{X}$ represents the sample matrix, $\bf{D}$, $\bf{{S}}$ are dictionary matrix and sparse codes matrix, respectively. $\bf{{W}}$ represents the learned classifier.}
	\label{fig:Flow_chat}
\end{figure}

Our framework contains two layers. Specifically, the first layer consists of the class specific dictionary learning for sparse representation (CSDL-SRC)~\cite{liu2016face} method, it is used to extract the crucial feature information of a class to wipe off singular points and improve robustness. The second layer is composed of LEDL which pulled the feature information belongs to different subspaces back into the same subspace to obtain the relationship among different classes. To be attention, different from the recent work CDPL-Net~\cite{zhang2020convolutional} which is a deep dictionary learning network, our proposed framework is a whole structure that can replace the dictionary pair learning layer of the network.

In addition, the feature after the first layer has been a better representation as to the reason that the objective function introduced the discriminative information and map them into different subspaces for the classifier which trained from the second layer. Thus, we can regard the first layer as a helpful tool to obtain excellent features. It does not work if we set the class shared dictionary learning method as the first layer and the class specific dictionary learning as the second layer. Figure~\ref{fig:Different_features} shows the variation of sample distribution. Figure~\ref{fig:Different_features} $a$ shows the random distribution samples belong to three classes; Figure~\ref{fig:Different_features} $b$ shows that the samples belong to the same class are clustered while the samples of three classes are in different subspaces; Figure~\ref{fig:Different_features} $c$ shows that the samples in different subspaces are pulled back into the same subspace. A schematic description of our proposed CDLF is given in Figure~\ref{fig:Flow_chat}. 

We adopt the alternating direction method of multipliers (ADMM)~\cite{boyd2011distributed} algorithm and blockwise coordinate descent (BCD)~\cite{liu2014blockwise} algorithm to optimize CDLF.
The contributions of this work are four-fold:
\begin{itemize}
\item We propose a novel class shared dictionary learning method named label embedded dictionary learning (LEDL) that introduces the $\ell_1$-norm regularization term as the sparse constraint. The $\ell_1$-norm sparse constraint can easily find the optimal sparse solution.
\item We propose a novel dictionary learning framework named cascaded dictionary learning framework (CDLF), which is the first time we combine the class specific and shared dictionary learning. More specifically, discriminative information is used in different ways to fully describe the feature while completely maintain the discriminative information. The CDLF can be considered as the extension of conventional dictionary learning algorithms.
\item We propose to utilize the alternating direction method of multipliers (ADMM)~\cite{boyd2011distributed} algorithm and blockwise coordinate descent (BCD)~\cite{liu2014blockwise} algorithm to optimize each layer of a dictionary learning task.
\item We evaluate the proposed LEDL and CDLF methods on six benchmark datasets, and the achieved superior performance verifies the effectiveness of our proposed methods.
\end{itemize}

We organize the rest of the paper as follows. Section ~\ref{Related work} introduces two related algorithms, which are CSDL-SRC and LC-KSVD. Section~\ref{Methodology} presents CDLF for image classification. section~\ref{Experimental results} shows some experiments and analyses. Finally, we conclude this paper in Section~\ref{Conclusion}.

\section{Related Work}
\label{Related work}
In this section, we overview two related dictionary learning methods, including class specific dictionary learning for sparse representation (CSDL-SRC) and label consistent K-SVD (LC-KSVD).

\subsection{Class specific dictionary learning for sparse representation (CSDL-SRC)}
\label{CSDL-SRC}

Liu \emph{et al.}~\cite{liu2016face} proposed CSDL-SRC to reduce the high residual error and instability of SRC. The authors consider the weight of each sample feature when generating the dictionary. Assume that \begin{scriptsize}${\bf{X}} =  \left[ {{{\bf{X}}^1},{{\bf{X}}^2}, \cdots ,{{\bf{X}}^C}} \right]  \in {{\mathbb{R}}^{d\times N}}$\end{scriptsize} is the training sample matrix, where $d$ represents the dimensions of the sample features, \begin{scriptsize}$N$\end{scriptsize} and \begin{scriptsize}$C$\end{scriptsize} are the number of training samples and the class number of training samples, respectively. The  \begin{scriptsize}$c_{th}$\end{scriptsize} class of training sample matrix is denoted as \begin{scriptsize}${{\bf{X}}^c}\in {{\mathbb{R}}^{d\times N^c}}$\end{scriptsize}, where \begin{scriptsize}$c = 1,2, \cdots ,C$\end{scriptsize} and \begin{scriptsize}$N^c$\end{scriptsize} is the \begin{scriptsize}$c_{th}$\end{scriptsize} class of \begin{scriptsize}$N$\end{scriptsize}(\begin{scriptsize}$N = \sum\limits_{c = 1}^C {{N^c}} $\end{scriptsize}). Liu \emph{et al.} build a weight coefficient matrix \begin{scriptsize}${{\bf{P}}^c}\in {{\mathbb{R}}^{{N^c}\times {K^c}}}$\end{scriptsize} for \begin{scriptsize}${{\bf{X}}^c}$\end{scriptsize}, where \begin{scriptsize}${K}$\end{scriptsize} is the dictionary size of CSDL-SRC and \begin{scriptsize}${K^c}$\end{scriptsize} is the \begin{scriptsize}$c_{th}$\end{scriptsize} class of \begin{scriptsize}$K$\end{scriptsize} (\begin{scriptsize}${K} = \sum\limits_{c = 1}^C {K^c} $\end{scriptsize}). The objective function of CSDL-SRC is as follows:
\begin{equation}
\scriptsize
\begin{split}
< {{\bf{P}}^c},{{\bf{U}}^c} >  &= \mathop {\arg \min }\limits_{{{\bf{P}}^c},{{\bf{U}}^c}} \left\| {{{\bf{X}}^c} - {{\bf{X}}^c}{{\bf{P}}^c}{{\bf{U}}^c}} \right\|_F^2 + 2\zeta {\left\| {{{\bf{U}}^c}} \right\|_{\ell_1}}\\
&s.t.\left\| {\bf{X}^c{\bf{P}}_{ \bullet k}^c} \right\|_2^2 \le 1\;\;\;(k = 1,2, \cdots, {K})
\end{split}\label{CSDL}
\end{equation}
where \begin{scriptsize}${{\bf{U}}^c}\in {{\mathbb{R}}^{{K^c}\times N^c}}$\end{scriptsize} is the sparse codes of \begin{scriptsize}${\bf{X}^c}$\end{scriptsize}, the \begin{scriptsize}$\ell_1$\end{scriptsize}-norm regularization term is utilized to enforce the sparsity, \begin{scriptsize}$\zeta$\end{scriptsize} is the regularization parameter to control the tradeoff between fitting goodness and sparseness. The \begin{scriptsize}${ {\left(  \bullet  \right)} _{ \bullet k}}$\end{scriptsize} denote the \begin{scriptsize}$k_{th}$\end{scriptsize} column vector of matrix  \begin{scriptsize}${ \left(  \bullet  \right) }$\end{scriptsize}.

\subsection{Label consistent K-SVD (LC-KSVD)}
\label{LC-KSVD}

Jiang \emph{et al.}~\cite{jiang2013label} proposed LC-KSVD to combine the discriminative sparse codes error with the reconstruction error and the classification error to form a unified objective function which is defined as follows:
\begin{equation}
\scriptsize
\begin{split}
< {\bf{B}},{\bf{W}},{\bf{A}},{\bf{V}} > &= \mathop {\arg \min }\limits_{{\bf{B}},{\bf{W}},{\bf{A}},{\bf{V}}} \left\| {{\bf{X}} - {\bf{BV}}} \right\|_F^2 + \lambda \left\| {{\bf{H}} - {\bf{WV}}} \right\|_F^2 + \omega \left\| {{\bf{Q}} - {\bf{AV}}} \right\|_F^2
\\{\kern 12pt} & s.t.{\kern 3pt} {\left\| {{{\bf{v}}_i}} \right\|_0} < T {\kern 6pt} \left( {i = 1,2 \cdots ,N} \right)
\end{split}\label{LC-KSVD}
\end{equation}
where \begin{scriptsize}$T$\end{scriptsize} is the sparsity constraint factor, \begin{scriptsize}${\bf{B}} \in {{\mathbb{R}}^{d\times K}}$\end{scriptsize} is the dictionary matrix of \begin{scriptsize}$\bf{X}$\end{scriptsize}, \begin{scriptsize}${\bf{V}} \in {{\mathbb{R}}^{K\times d}}$\end{scriptsize} is the sparse codes matrix of \begin{scriptsize}${\bf{X}}$\end{scriptsize}. \begin{scriptsize}${\bf{W}} \in {{\mathbb{R}}^{C\times K}}$\end{scriptsize} is a classifier matrix learned from the given label matrix \begin{scriptsize}${\bf{H}} \in {{\mathbb{R}}^{C\times N}}$\end{scriptsize}. We hope \begin{scriptsize}${\bf{W}}$\end{scriptsize} can return the most probable class this sample belongs to. \begin{scriptsize}${\bf{Q}} \in {{\mathbb{R}}^{K\times N}}$\end{scriptsize} represents the discriminative sparse codes matrix and \begin{scriptsize}${\bf{A}} = \left[ {{{\bf{a}}_1},{{\bf{a}}_2}, \cdots ,{{\bf{a}}_{K_2}}} \right] \in {{\mathbb{R}}^{K_2\times K_2}}$\end{scriptsize} is a linear transformation matrix relys on \begin{scriptsize}${\bf{Q}}$\end{scriptsize}. \begin{scriptsize}$\lambda$\end{scriptsize} and \begin{scriptsize}$\omega$\end{scriptsize} are the regularization parameters balancing the discriminative sparse codes errors and the classification contribution to the overall objective function, respectively.

\section{Methodology}
\label{Methodology}
In this section, we elaborate on the construction of a cascaded dictionary learning framework (CDLF). Specifically, in subsection~\ref{ledl}, we introduce the label embedded dictionary learning method. In subsection~\ref{CDLF}, we propose the cascaded method of the first layer and the second layer. In subsection~\ref{Optimization}, we give the optimization method of objective function. 

\subsection{Label embedded dictionary learning (LEDL)}
\label{ledl}

This subsection proposes a novel dictionary learning method named label embedded dictionary learning (LEDL) for image classification. LEDL is an improvement of LC-KSVD, which introducing the $\ell_1$-norm regularization term to replace the $\ell_0$-norm regularization of LC-KSVD. Thus, we can freely select the basis vectors for linear fitting to obtain an optimal sparse solution instead of a suboptimal solution.

The objection function is as follows and the $\varepsilon$ is the regularization parameter:

\begin{equation}
    \scriptsize
	\begin{split}
		< {\bf{B}},{\bf{W}},{\bf{A}},{\bf{V}} > = &\mathop {\arg \min }\limits_{{\bf{B}},{\bf{W}},{\bf{A}},{\bf{V}}} \left\| {{\bf{X}} - {\bf{BV}}} \right\|_F^2 + \lambda \left\| {{\bf{H}} - {\bf{WV}}} \right\|_F^2 \\&+ \omega \left\| {{\bf{Q}} - {\bf{AV}}} \right\|_F^2 + 2\varepsilon {\left\| {\bf{V}} \right\|_{\ell_1}}\\
		{\rm{s}}.t.{\kern 4pt}\left\| {{{\bf{B}}_{ \bullet k}}} \right\|_2^2 \le 1, {\kern 1pt} {\kern 1pt} &\left\| {{{\bf{W}}_{ \bullet k}}} \right\|_2^2 \le 1,{\kern 1pt}
		\left\| {{{\bf{A}}_{ \bullet k}}} \right\|_2^2 \le 1{\kern 3pt}\left( {k = 1,2, \cdots K} \right)
	\end{split}\label{LEDL}
\end{equation}
Where the definitions of \begin{scriptsize}$\bf{W}$\end{scriptsize}, \begin{scriptsize}$\bf{W}$\end{scriptsize}, \begin{scriptsize}$\bf{A}$\end{scriptsize}, \begin{scriptsize}$\bf{V}$\end{scriptsize} in Equation~\ref{LEDL} are same with the ones in Equation~\ref{LC-KSVD}. Consider the optimization problem (\ref{LEDL}) is not jointly convex in both \begin{scriptsize}${\bf{V}}$\end{scriptsize}, \begin{scriptsize}${\bf{B}}$\end{scriptsize}, \begin{scriptsize}${\bf{W}}$\end{scriptsize} and \begin{scriptsize}${\bf{A}}$\end{scriptsize}, it is separately convex in either \begin{scriptsize}${\bf{V}}$\end{scriptsize} (with \begin{scriptsize}${\bf{B}}$\end{scriptsize}, \begin{scriptsize}${\bf{W}}$\end{scriptsize}, \begin{scriptsize}${\bf{A}}$\end{scriptsize} fixed), \begin{scriptsize}${\bf{B}}$\end{scriptsize} (with \begin{scriptsize}${\bf{V}}$\end{scriptsize}, \begin{scriptsize}${\bf{W}}$\end{scriptsize}, \begin{scriptsize}${\bf{A}}$\end{scriptsize} fixed), \begin{scriptsize}${\bf{W}}$\end{scriptsize} (with \begin{scriptsize}${\bf{V}}$\end{scriptsize}, \begin{scriptsize}${\bf{B}}$\end{scriptsize}, \begin{scriptsize}${\bf{A}}$\end{scriptsize} fixed) or \begin{scriptsize}${\bf{A}}$\end{scriptsize} (with \begin{scriptsize}${\bf{V}}$\end{scriptsize}, \begin{scriptsize}${\bf{B}}$\end{scriptsize}, \begin{scriptsize}${\bf{W}}$\end{scriptsize} fixed). Thus, we recognised the optimization as four optimization subproblems. Futhermore, ADMM~\cite{boyd2011distributed} framework is employed to finding sparse codes (\begin{scriptsize}$\bf{V}$\end{scriptsize}) and BCD~\cite{liu2014blockwise} algorithm is utilised to getting the learning bases (\begin{scriptsize}$\bf{B}$\end{scriptsize}, \begin{scriptsize}$\bf{W}$\end{scriptsize}, \begin{scriptsize}$\bf{A}$\end{scriptsize}). The details of optimization are parts of CDLF, which are shown in next section~\ref{CDLF}.    

\subsection{Cascaded dictionary learning framework (CDLF)}
\label{CDLF}

In this subsection, we introduce the first layer of the framework which is composed of CSDL-SRC, then we rewrite the objective function of the proposed LEDL and let it be the second layer.

\subsubsection{The first layer}
\label{first layer}

The first layer is utilised to learn a class specific dictionary for each class. Given a training sample matrix \begin{scriptsize}$\bf{X}$\end{scriptsize}, then we set a suitable dictionary size \begin{scriptsize}$K_1$\end{scriptsize}, the objective function of the first layer is as follows:
\begin{equation}
\scriptsize
\begin{split}
< {\bf{D}}_{{l_1}}^c,{\bf{S}}_{{l_1}}^c &>  = \mathop {\arg \min }\limits_{{\bf{D}}_{{l_1}}^c,{\bf{S}}_{{l_1}}^c} \left\| {{{\bf{X}}^c} - {\bf{D}}_{{l_1}}^c{\bf{S}}_{{l_1}}^c} \right\|_F^2 + 2\zeta {\left\| {{\bf{S}}_{{l_1}}^c} \right\|_{\ell_1}}\\
&s.t.\left\| {{{\left( {{\bf{D}}_{{l_1}}^c} \right)}_{ \bullet k}}} \right\|_2^2 \le 1\;\;\;(k = 1,2, \cdots, {K_1})
\end{split}\label{CSDL1}
\end{equation}
where \begin{scriptsize}${{\bf{D}}_{{l_1}}} \in {{\mathbb{R}}^{{d}\times K_1}}$\end{scriptsize} and \begin{scriptsize}${{\bf{S}}_{{l_1}}} \in {{\mathbb{R}}^{{K_1}\times N}}$\end{scriptsize} are the dictionary matrix and sparse codes matrix of the first layer in our proposed CDLF, respectively. 

\subsubsection{The second layer}
\label{second layer}

The second layer is composed of the proposed LEDL, which is used to learn a class shared dictionary. Based on the computation above, we explicitly construct a sparse codes matrix \begin{scriptsize}${{\bf{S}}_{{l_1}}}$\end{scriptsize} from the first layer and make it to be one of the input of the next layer. In addition, the label matrix \begin{scriptsize}${{\bf{H}}} \in {{\mathbb{R}}^{{C}\times N}}$\end{scriptsize} and discriminative sparse codes matrix \begin{scriptsize}${{\bf{Q}}} \in {{\mathbb{R}}^{{K_2}\times N}}$\end{scriptsize} are also introduced to the second layer. After giving a reasonable dictionary size \begin{scriptsize}$K_2$\end{scriptsize} of LEDL, the objective function can be rewritten as follows:
\begin{equation}
\scriptsize
\begin{split}
< {{\bf{D}}_{{l_2}}},&{\bf{W}},{\bf{A}},{{\bf{S}}_{{l_2}}} >  = \mathop {\arg \min }\limits_{{{\bf{D}}_{{l_2}}},{\bf{W}},{\bf{A}},{{\bf{S}}_{{l_2}}}} \left\| {{{\bf{S}}_{{l_1}}} - {{\bf{D}}_{{l_2}}}{{\bf{S}}_{{l_2}}}} \right\|_F^2 
\\&+ \lambda \left\| {{\bf{H}} - {\bf{W}}{{\bf{S}}_{{l_2}}}} \right\|_F^2 + \omega \left\| {{\bf{Q}} - {\bf{A}}{{\bf{S}}_{{l_2}}}} \right\|_F^2 + 2\varepsilon {\left\| {{{\bf{S}}_{{l_2}}}} \right\|_{\ell_1}}
\\&s.t.{\kern 1pt} {\kern 1pt} {\kern 1pt} {\kern 1pt} {\kern 1pt} \left\| {{{\left( {{{\bf{D}}_{{l_2}}}} \right)}_{ \bullet k}}} \right\|_2^2 \le 1,{\kern 1pt} {\kern 1pt} {\kern 1pt} \left\| {{{\bf{W}}_{ \bullet k}}} \right\|_2^2 \le 1,
\\&{\kern 17pt} {\kern 1pt} {\kern 1pt} \left\| {{{\bf{A}}_{ \bullet k}}} \right\|_2^2 \le 1{\kern 1pt} {\kern 1pt} {\kern 1pt} {\kern 1pt} {\kern 1pt} (k = 1,2 \cdots, {K_2})
\end{split}\label{LEDL1}
\end{equation}
where \begin{scriptsize}${{\bf{D}}_{{l_2}}} \in {{\mathbb{R}}^{K_1\times K_2}}$\end{scriptsize} is the dictionary of \begin{scriptsize}${{\bf{S}}_{{l_1}}}$, ${{\bf{S}}_{{l_2}}} \in {{\mathbb{R}}^{K_2\times N}}$\end{scriptsize} is the sparse codes of \begin{scriptsize}${{\bf{S}}_{{l_1}}}$\end{scriptsize}. The definitions of \begin{scriptsize}${{\bf{W}}} \in {{\mathbb{R}}^{C\times K_2}}$ and ${{\bf{A}}} \in {{\mathbb{R}}^{K_2 \times K_2}}$\end{scriptsize} in Equation~\ref{LEDL1} are the same with the ones in Equation~\ref{LC-KSVD}. 

\subsection{Optimization of objective function}
\label{Optimization}

Due to the optimization issues about Equation~\ref{CSDL1} and Equation~\ref{LEDL1} are not jointly convex, Equation~\ref{CSDL1} is separately convex in either \begin{scriptsize}${\bf{S}}_{{l_1}}^c$\end{scriptsize}(with \begin{scriptsize}${\bf{D}}_{{l_1}}^c$\end{scriptsize} fixed) or \begin{scriptsize}${\bf{D}}_{{l_1}}^c$\end{scriptsize}(with \begin{scriptsize}${\bf{S}}_{{l_1}}^c$\end{scriptsize} fixed), and Equation~\ref{LEDL1} is separately convex in either \begin{scriptsize}${{\bf{S}}_{{l_2}}}$\end{scriptsize}(with \begin{scriptsize}${{\bf{D}}_{{l_2}}}$\end{scriptsize}, \begin{scriptsize}${\bf{W}}$\end{scriptsize}, \begin{scriptsize}${\bf{A}}$\end{scriptsize} fixed), \begin{scriptsize}${{\bf{D}}_{{l_2}}}$\end{scriptsize}(with \begin{scriptsize}${\bf{W}}$\end{scriptsize}, \begin{scriptsize}${\bf{A}}$\end{scriptsize}, \begin{scriptsize}${{\bf{S}}_{{l_2}}}$\end{scriptsize} fixed), \begin{scriptsize}${\bf{W}}$\end{scriptsize}(with \begin{scriptsize}${{\bf{D}}_{{l_2}}}$\end{scriptsize}, \begin{scriptsize}${\bf{A}}$\end{scriptsize}, \begin{scriptsize}${{\bf{S}}_{{l_2}}}$\end{scriptsize} fixed), or \begin{scriptsize}${\bf{A}}$\end{scriptsize}(with \begin{scriptsize}${{\bf{D}}_{{l_2}}}$\end{scriptsize}, \begin{scriptsize}${\bf{W}}$, ${{\bf{S}}_{{l_2}}}$\end{scriptsize} fixed). To this end, we cast the optimization problem as six subproblems which are \begin{scriptsize}$\ell_1$\end{scriptsize}-norm regularized least-squares(\begin{scriptsize}$\ell_1$\end{scriptsize}-\begin{scriptsize}$\ell_s$\end{scriptsize}) minimization subproblem for finding sparse codes(\begin{scriptsize}${{\bf{S}}_{{l_1}}^c}$\end{scriptsize}, \begin{scriptsize}${{\bf{S}}_{{l_2}}}$\end{scriptsize}) and \begin{scriptsize}$\ell_1$\end{scriptsize}-norm constrained least-squares (\begin{scriptsize}$\ell_1$\end{scriptsize}-\begin{scriptsize}$\ell_s$\end{scriptsize}) minimization subproblem for learning bases (\begin{scriptsize}${{\bf{D}}_{{l_1}}^c}$\end{scriptsize}, \begin{scriptsize}${{\bf{D}}_{{l_2}}}$\end{scriptsize}, \begin{scriptsize}${{\bf{W}}}$\end{scriptsize}, \begin{scriptsize}${{\bf{A}}}$\end{scriptsize}), respectively. Here, ADMM~\cite{boyd2011distributed} framework is introduced to solve the first subproblem while BCD~\cite{liu2014blockwise} method offers the key to addressing the other subproblems. 

\subsubsection{Optimization of the first layer}

ADMM is usually used to solve the equality-constrained problem while the objective function of CSDL-SRC is unconstrained. Thus the core idea of imposing ADMM framework here is to introduce an auxiliary variable to reformulate the original function into a linear equality-constrained problem. By introducing the auxiliary variable \begin{scriptsize}${\bf{Z}}_{{l_1}}^c$\end{scriptsize}, the \begin{scriptsize}${{{\bf{S}}_{{l_1}}}}$\end{scriptsize} in Equation~\ref{CSDL1} can be substituted by \begin{scriptsize}${{{\bf{C}}_{{l_1}}}}$\end{scriptsize} and \begin{scriptsize}${{{\bf{Z}}_{{l_1}}}}$\end{scriptsize}, thus we can rewritten Equation~\ref{CSDL1} as follows:
\begin{equation}
\scriptsize
\begin{split}
< {\bf{D}}_{{l_1}}^c,{\bf{C}}_{{l_1}}^c,{\bf{Z}}_{{l_1}}^c >  \!\!\!\!&\!\!\!\!=\!\!\!\! \mathop {\arg \min }\limits_{{\bf{D}}_{{l_1}}^c,{\bf{C}}_{{l_1}}^c,{\bf{Z}}_{{l_1}}^c} \left\| {{{\bf{X}}^c} - {\bf{D}}_{{l_1}}^c{\bf{C}}_{{l_1}}^c} \right\|_F^2 + 2\zeta {\left\| {{\bf{Z}}_{{l_1}}^c} \right\|_{\ell_1}}\!\!\!\!
\\s.t.\;\;{\bf{C}}_{{l_1}}^c = {\bf{Z}}_{{l_1}}^c,\;\;
&\left\| {{{\left( {{\bf{D}}_{{l_1}}^c} \right)}_{ \bullet k}}} \right\|_2^2 \le 1\;\;\;(k = 1,2, \cdots, {K_1})
\end{split}\label{CSDL2}
\end{equation}

Then the lagrangian function of the problem (\ref{CSDL2}) with fixed \begin{scriptsize}${\bf{D}}_{{l_1}}^c$\end{scriptsize} can be rewritten as: 

\begin{equation}
\scriptsize
\begin{split}
< {\bf{C}}_{{l_1}}^c,{\bf{Z}}_{{l_1}}^c,{\bf{L}}_{{l_1}}^c >  &= \mathop {\arg \min }\limits_{{\bf{C}}_{{l_1}}^c,{\bf{Z}}_{{l_1}}^c,{\bf{L}}_{{l_1}}^c} \left\| {{{\bf{X}}^c} - {\bf{D}}_{{l_1}}^c{\bf{C}}_{{l_1}}^c} \right\|_F^2 
\\&+ 2\zeta {\left\| {{\bf{Z}}_{{l_1}}^c} \right\|_{\ell_1}} + 2{\left( {{\bf{L}}_{{l_1}}^c} \right)^T}\left( {{\bf{C}}_{{l_1}}^c - {\bf{Z}}_{{l_1}}^c} \right) 
\\&+ \varphi \left\| {{\bf{C}}_{{l_1}}^c - {\bf{Z}}_{{l_1}}^c} \right\|_F^2
\end{split}\label{CSDL3}
\end{equation}

where \begin{scriptsize}${\bf{L}}_{{l_1}}^c \in {{\mathbb{R}}^{{K_1^c} \times {N^c}}}$\end{scriptsize} is the augmented lagrangian multiplier and \begin{scriptsize}$ \varphi>0$\end{scriptsize} is the penalty parameter. We can gain the closed-form solution with respect to each iteration by follows:
\\\\(1) {\bfseries Updating \begin{scriptsize}${\bf{C}}_{{l_1}}^c$\end{scriptsize} while fixing \begin{scriptsize}${\bf{D}}_{{l_1}}^c$, ${\bf{Z}}_{{l_1}}^c$\end{scriptsize} and \begin{scriptsize}${{\bf{L}}_{{l_1}}^c}$\end{scriptsize}}:
\begin{equation}
\scriptsize
\begin{split}{\left( {{\bf{C}}_{{l_1}}^c} \right)_{m + 1}} =  < {\left( {{\bf{D}}_{{l_1}}^c} \right)_m},{\left( {{\bf{C}}_{{l_1}}^c} \right)_m},{\left( {{\bf{Z}}_{{l_1}}^c} \right)_m},{\left( {{\bf{L}}_{{l_1}}^c} \right)_m} > 
\end{split}\label{Update C_{l1}_1}
\end{equation}
where \begin{scriptsize}$m\left( {m = 0,1,2, \cdots } \right)$\end{scriptsize} is the iteration number and \begin{scriptsize}${\left(  \bullet  \right)_m}$\end{scriptsize} means the value of matrix \begin{scriptsize}${ \left(  \bullet  \right) }$\end{scriptsize} after \begin{scriptsize}${m_{th}}$\end{scriptsize} iteration, the closed form solution of \begin{scriptsize}${{\bf{C}}_{{l_1}}^c}$\end{scriptsize} is:

\begin{equation}
\scriptsize
\begin{split}
{\left( {{\bf{C}}_{{l_1}}^c} \right)_{m + 1}} = {\left( {{\bf{\dot C}}_{{l_1}}^c} \right)^{ - 1}}{\bf{\ddot C}}_{{l_1}}^c
\end{split}\label{Update C_{l1}_2}
\end{equation} 	

the \begin{scriptsize}${\bf{\dot C}}_{{l_1}}^c $\end{scriptsize} here can be written as:
\begin{equation}
\scriptsize
\begin{split}
{\bf{\dot C}}_{{l_1}}^c = \left( {{\bf{D}}_{{l_1}}^c} \right)_m^T{\left( {{\bf{D}}_{{l_1}}^c} \right)_m} + \varphi {\bf{I}}
\end{split}\label{Update C_{l1}_3}
\end{equation} 

where \begin{scriptsize}$\bf{I}$\end{scriptsize} is the identity matrix. The \begin{scriptsize}${\bf{\ddot C}}_{{l_1}}^c$\end{scriptsize} here can be written as:
\begin{equation}
\scriptsize
\begin{split}
{\bf{\ddot C}}_{{l_1}}^c = \left( {{\bf{D}}_{{l_1}}^c} \right)_m^T{{\bf{X}}^c} - {\left( {{\bf{L}}_{{l_1}}^c} \right)_m} + \varphi {\left( {{\bf{Z}}_{{l_1}}^c} \right)_m}
\end{split}\label{Update C_{l1}_4}
\end{equation} 
\\\\(2) {\bfseries Updating \begin{scriptsize}${\bf{Z}}_{{l_1}}^c$\end{scriptsize} while fixing \begin{scriptsize}${\bf{D}}_{{l_1}}^c$\end{scriptsize}, \begin{scriptsize}${\bf{C}}_{{l_1}}^c$\end{scriptsize} and \begin{scriptsize}${{\bf{L}}_{{l_1}}^c}$\end{scriptsize}}:

\begin{equation}
\scriptsize
\begin{split}
{\left( {{\bf{Z}}_{{l_1}}^c} \right)_{m + 1}} =  < {\left( {{\bf{D}}_{{l_1}}^c} \right)_m},{\left( {{\bf{C}}_{{l_1}}^c} \right)_{m{\rm{ + 1}}}},{\left( {{\bf{Z}}_{{l_1}}^c} \right)_m},{\left( {{\bf{L}}_{{l_1}}^c} \right)_m} >
\end{split}\label{Update Z_{l1}_1}
\end{equation}

the closed form solution of \begin{scriptsize}${\bf{Z}}_{{l_1}}^c$\end{scriptsize} is:
\begin{equation}
\scriptsize
\begin{split}
{\left( {{\bf{Z}}_{{l_1}}^c} \right)_{m + 1}} = {\bf{\dot Z}}_{{l_1}}^c + {\bf{\ddot Z}}_{{l_1}}^c
\end{split}\label{Update Z_{l1}_2}
\end{equation}

the \begin{scriptsize}${\bf{\dot Z}}_{{l_1}}^c$\end{scriptsize} here can be written as:

\begin{equation}
\scriptsize
\begin{split}
{\bf{\dot Z}}_{{l_1}}^c = \max \left\{ {{{\left( {{\bf{C}}_{{l_1}}^c} \right)}_{m + 1}} + \frac{{{{\left( {{\bf{L}}_{{l_1}}^c} \right)}_m}}}{\varphi } - \frac{\zeta }{\varphi }{\bf{I}},{\bf{0}}} \right\}
\end{split}\label{Update Z_{l1}_3}
\end{equation}

the \begin{scriptsize}${\bf{\ddot Z}}_{{l_1}}^c$\end{scriptsize} here can be written as:
\begin{equation}
\scriptsize
\begin{split}
{\bf{\ddot Z}}_{{l_1}}^c = \min \left\{ {{{\left( {{\bf{C}}_{{l_1}}^c} \right)}_{m + 1}} + \frac{{{{\left( {{\bf{L}}_{{l_1}}^c} \right)}_m}}}{\varphi } + \frac{\zeta }{\varphi }{\bf{I}},{\bf{0}}} \right\}
\end{split}\label{Update Z_{l1}_4}
\end{equation}
\\(3) {\bfseries Updating \begin{scriptsize}${{\bf{L}}_{{l_1}}^c}$\end{scriptsize} while fixing \begin{scriptsize}${\bf{D}}_{{l_1}}^c$, ${\bf{C}}_{{l_1}}^c$\end{scriptsize} and \begin{scriptsize}${\bf{Z}}_{{l_1}}^c$\end{scriptsize}}:
\begin{equation}
\scriptsize
\begin{split}
{\left( {{\bf{L}}_{{l_1}}^c} \right)_{m + 1}} = {\left( {{\bf{L}}_{{l_1}}^c} \right)_m} + \varphi  \left( {{{\left( {{\bf{C}}_{{l_1}}^c} \right)}_{m + 1}} - {{\left( {{\bf{Z}}_{{l_1}}^c} \right)}_{m + 1}}} \right)
\end{split}\label{Update L_{l1}_1}
\end{equation}

Based on the above ADMM steps, we obtain the closed form solution of \begin{scriptsize}${\bf{C}}_{{l_1}}^c$, ${\bf{Z}}_{{l_1}}^c$\end{scriptsize} and \begin{scriptsize}${{\bf{L}}_{{l_1}}^c}$\end{scriptsize}.
Then we utilise BCD method with fixed \begin{scriptsize}${\bf{C}}_{{l_1}}^c$\end{scriptsize}, \begin{scriptsize}${\bf{Z}}_{{l_1}}^c$\end{scriptsize} and \begin{scriptsize}${{\bf{L}}_{{l_1}}^c}$\end{scriptsize} to solve the constrained minimization problem of Equation~\ref{CSDL2}. The objective function can be rewritten as follows:
\begin{equation}
\scriptsize
\begin{split}
< {\bf{D}}_{{l_1}}^c >  = \mathop {\arg \min }\limits_{{\bf{D}}_{{l_1}}^c} \left\| {{{\bf{X}}^c} - {\bf{D}}_{{l_1}}^c{\bf{C}}_{{l_1}}^c} \right\|_F^2 + 2\zeta {\left\| {{\bf{Z}}_{{l_1}}^c} \right\|_{\ell_1}}\\
+ 2{\left( {{\bf{L}}_{{l_1}}^c} \right)^T}\left( {{\bf{C}}_{{l_1}}^c - {\bf{Z}}_{{l_1}}^c} \right) + \varphi \left\| {{\bf{C}}_{{l_1}}^c - {\bf{Z}}_{{l_1}}^c} \right\|_F^2\\
s.t.\;\;\;\left\| {{{\left( {{\bf{D}}_{{l_1}}^c} \right)}_{ \bullet k}}} \right\|_2^2 \le 1\;\;\;(k = 1,2, \cdots, {K_1})
\end{split}\label{CSDL4}
\end{equation}

To this end, we can solve the closed-form solution with respect to the single column by follows:
\\\\(4) {\bfseries Updating \begin{scriptsize}${\bf{D}}_{{l_1}}^c$\end{scriptsize} while fixing \begin{scriptsize}${\bf{C}}_{{l_1}}^c$\end{scriptsize}, \begin{scriptsize}${\bf{Z}}_{{l_1}}^c$\end{scriptsize} and ${{\bf{L}}_{{l_1}}^c}$}:
\begin{equation}
\scriptsize
\begin{split}
{\left( {{\bf{D}}_{{l_1}}^c} \right)_{m + 1}} \!\!\!\!=  < {\left( {{\bf{D}}_{{l_1}}^c} \right)_m},{\left( {{\bf{C}}_{{l_1}}^c} \right)_{m{\rm{ + 1}}}},{\left( {{\bf{Z}}_{{l_1}}^c} \right)_{m + 1}},{\left( {{\bf{L}}_{{l_1}}^c} \right)_{m + 1}} >
\end{split}\label{Update D_{l1}_1}
\end{equation}
the closed form solution of ${\bf{D}}_{{l_1}}^c$ is:
\begin{equation}
\scriptsize
\begin{split}
\left( {{{\left( {{\bf{D}}_{{l_1}}^c} \right)}_{ \bullet k}}} \right){{\kern 1pt} _{m + 1}}{\kern 3pt}{\rm{ = }}{\kern 3pt}\frac{{{\bf{\dot D}}_{{l_1}}^c}}{{{{\left\| {{\bf{\dot D}}_{{l_1}}^c} \right\|}_2}}}
\end{split}\label{Update D_{l1}_2}
\end{equation}
the \begin{scriptsize}${{\bf{\dot D}}_{{l_1}}^c}$\end{scriptsize} here can be written as:
\begin{equation}
\scriptsize
\begin{split}
{\bf{\dot D}}_{{l_1}}^c&{\kern 3pt}{\rm{ = }}{\kern 3pt}{{\bf{X}}^c}{\left[ {{{\left( {{{\left( {{\bf{C}}_{{l_1}}^c} \right)}_{k \bullet }}} \right)}_{m + 1}}} \right]^T}
- {\left( {{{\left( {{\bf{\tilde D}}_{{l_1}}^c} \right)}^k}} \right)_m}{\left( {{\bf{C}}_{{l_1}}^c} \right)_{m + 1}}{\left[ {{{\left( {{{\left( {{\bf{C}}_{{l_1}}^c} \right)}_{k \bullet }}} \right)}_{m + 1}}} \right]^T}
\end{split}\label{Update D_{l1}_3}
\end{equation}
where \begin{scriptsize}${\left( {{\bf{\tilde D}}_{{l_1}}^c} \right)^k} = \left\{ {\begin{array}{*{20}{c}}
		{{{\left( {{\bf{D}}_{{l_1}}^c} \right)}_{ \bullet p}},{\kern 1pt} {\kern 1pt} {\kern 1pt} {\kern 1pt} {\kern 1pt} {\kern 1pt} {\kern 1pt} {\kern 1pt} p \ne k}\\
		{{\kern 1pt} {\kern 1pt} {\kern 1pt} {\kern 1pt} {\kern 1pt} {\kern 1pt} {\bf{0}},{\kern 1pt} {\kern 1pt} {\kern 1pt} {\kern 1pt} {\kern 1pt} {\kern 1pt} {\kern 1pt} {\kern 1pt} {\kern 1pt} {\kern 1pt} {\kern 1pt} {\kern 1pt} {\kern 1pt} {\kern 1pt} {\kern 1pt} {\kern 1pt} {\kern 1pt} {\kern 1pt} {\kern 1pt} {\kern 1pt} {\kern 1pt} {\kern 1pt} {\kern 1pt} {\kern 1pt} {\kern 1pt} {\kern 1pt} {\kern 1pt} {\kern 1pt} {\kern 1pt} {\kern 1pt} p = k}
		\end{array}} \right.$, ${\left(  \bullet  \right)_{k \bullet }}$\end{scriptsize} denote the \begin{scriptsize}$k_{th}$\end{scriptsize} row vector of matrix  \begin{scriptsize}${ \left(  \bullet  \right) }$\end{scriptsize}.

\subsubsection{Optimization of the second layer}

Like the above procedure, the LEDL problem can be decomposed into two subproblems, which are the same as those of CSDL-SRC, which can be optimized by ADMM and BCD methods, respectively. 

For finding sparse codes subproblem, we utilise AD-MM method to optimize the objective function, hence the Equation~\ref{LEDL1} with \begin{scriptsize}${{\bf{D}}_{{l_2}}}$, ${\bf{W}}$\end{scriptsize}, \begin{scriptsize}${\bf{A}}$\end{scriptsize} fixed can be written as follows:
\begin{equation}
\scriptsize
\begin{split}
< {{\bf{C}}_{{l_2}}},&{{\bf{Z}}_{{l_2}}},{{\bf{L}}_{{l_2}}} >  = \mathop {\arg \min }\limits_{{{\bf{C}}_{{l_2}}},{{\bf{Z}}_{{l_2}}},{{\bf{L}}_{{l_2}}}} \left\| {{{\bf{C}}_{{l_1}}} - {{\bf{D}}_{{l_2}}}{{\bf{C}}_{{l_2}}}} \right\|_F^2\\&
+ \lambda \left\| {{\bf{H}} - {\bf{W}}{{\bf{C}}_{{l_2}}}} \right\|_F^2 + \omega \left\| {{\bf{Q}} - {\bf{A}}{{\bf{C}}_{{l_2}}}} \right\|_F^2\\&
+ 2{\left( {{{\bf{L}}_{{l_2}}}} \right)^T}\left( {{{\bf{C}}_{{l_2}}} - {{\bf{Z}}_{{l_2}}}} \right) + \rho \left\| {{{\bf{C}}_{{l_2}}} - {{\bf{Z}}_{{l_2}}}} \right\|_F^2\\&
+ 2\varepsilon {\left\| {{{\bf{Z}}_{{l_2}}}} \right\|_{\ell_1}}
\end{split}\label{LEDL3}
\end{equation}
where the definitions and applications of \begin{scriptsize} ${\bf{C}}_{{l_2}}^c$\end{scriptsize}, \begin{scriptsize}${\bf{Z}}_{{l_2}}^c$\end{scriptsize}, \begin{scriptsize}${\bf{L}}_{{l_2}}^c$\end{scriptsize} and \begin{scriptsize}$\rho$\end{scriptsize} in Equation~\ref{LEDL3} are similar to the \begin{scriptsize}${\bf{C}}_{{l_1}}^c$, ${\bf{Z}}_{{l_1}}^c$, ${\bf{L}}_{{l_1}}^c$\end{scriptsize} and \begin{scriptsize}$\varphi$\end{scriptsize} in Equation~\ref{CSDL3}. Thus, we can obtain the closed-form solution with respect to each iteration by follows:
\\\\(1) {\bfseries Updating \begin{scriptsize}${\bf{C}}_{{l_2}}^c$\end{scriptsize} while fixing \begin{scriptsize}${\bf{D}}_{{l_2}}^c$\end{scriptsize}, \begin{scriptsize}${\bf{W}}$\end{scriptsize}, ${\bf{A}}$, \begin{scriptsize}${\bf{Z}}_{{l_2}}^c$\end{scriptsize} and \begin{scriptsize}${{\bf{L}}_{{l_2}}^c}$\end{scriptsize}, the closed-form solution of \begin{scriptsize}${\bf{C}}_{{l_2}}^c$\end{scriptsize} is}:
\begin{equation}
\scriptsize
\begin{split}
{\left( {{{\bf{C}}_{{l_2}}}} \right)_{m + 1}} = {\left( {{{{\bf{\dot C}}}_{{l_2}}}} \right)^{ - 1}}{{{\bf{\ddot C}}}_{{l_2}}}
\end{split}\label{Update C_{l2}_2}
\end{equation}
where 

\begin{equation}
\scriptsize
\begin{split}
{{{\bf{\dot C}}}_{{l_2}}} &= {\left( {{{\left( {{{\bf{D}}_{{l_2}}}} \right)}_m}} \right)^T}{\left( {{{\bf{D}}_{{l_2}}}} \right)_m} + \lambda {{\bf{W}}_m}^T{{\bf{W}}_m} 
+ \omega {{\bf{A}}_m}^T{{\bf{A}}_m} + \rho {\bf{I}}
\end{split}\label{Update C_{l2}_3}
\end{equation}

\begin{equation}
\scriptsize
\begin{split}
{{{\bf{\ddot C}}}_{{l_2}}} &= {\left( {{{\left( {{{\bf{D}}_{{l_2}}}} \right)}_m}} \right)^T}{\left( {{{\bf{C}}_{{l_1}}}} \right)_m} + \lambda {{\bf{W}}_m}^T{{\bf{H}}_m} 
+ \omega {{\bf{A}}_m}^T{{\bf{Q}}_m} - {\left( {{{\bf{L}}_{{l_2}}}} \right)_m} + \rho {\left( {{{\bf{Z}}_{{l_2}}}} \right)_m}
\end{split}\label{Update C_{l2}_4}
\end{equation}
\\\\(2) {\bfseries Updating \begin{scriptsize}${\bf{Z}}_{{l_2}}^c$\end{scriptsize} while fixing \begin{scriptsize}${\bf{D}}_{{l_2}}^c$, ${\bf{W}}$\end{scriptsize}, \begin{scriptsize}${\bf{A}}$\end{scriptsize}, \begin{scriptsize}${\bf{C}}_{{l_2}}^c$\end{scriptsize} and \begin{scriptsize}${{\bf{L}}_{{l_2}}^c}$\end{scriptsize}, the closed-form solution of \begin{scriptsize}${\bf{Z}}_{{l_2}}^c$\end{scriptsize} is}:
\begin{equation}
\scriptsize
\begin{split}
{\left( {{{\bf{Z}}_{{l_2}}}} \right)_{m + 1}} = {{{\bf{\dot Z}}}_{{l_2}}} + {{{\bf{\ddot Z}}}_{{l_2}}}
\end{split}\label{Update Z_{l2}_2}
\end{equation}
where

\begin{equation}
\scriptsize
\begin{split}
{{{\bf{\dot Z}}}_{{l_2}}} = \max \left\{ {{{\left( {{{\bf{C}}_{{l_2}}}} \right)}_{m + 1}} + \frac{{{{\left( {{{\bf{L}}_{{l_2}}}} \right)}_m}}}{\rho } - \frac{\varepsilon }{\rho }{\bf{I}},{\bf{0}}} \right\}
\end{split}\label{Update Z_{l2}_3}
\end{equation}

\begin{equation}
\scriptsize
\begin{split}
{{{\bf{\ddot Z}}}_{{l_2}}} = \min \left\{ {{{\left( {{{\bf{C}}_{{l_2}}}} \right)}_{m + 1}} + \frac{{{{\left( {{{\bf{L}}_{{l_2}}}} \right)}_m}}}{\rho } + \frac{\varepsilon }{\rho }{\bf{I}},{\bf{0}}} \right\}
\end{split}\label{Update Z_{l2}_4}
\end{equation}
\\\\(3) {\bfseries Updating \begin{scriptsize}${{\bf{L}}_{{l_2}}^c}$\end{scriptsize} while fixing \begin{scriptsize}${\bf{D}}_{{l_2}}^c$\end{scriptsize}, \begin{scriptsize}${\bf{W}}$\end{scriptsize}, \begin{scriptsize}${\bf{A}}$\end{scriptsize}, \begin{scriptsize}${\bf{C}}_{{l_2}}^c$\end{scriptsize} and \begin{scriptsize}${\bf{Z}}_{{l_2}}^c$\end{scriptsize}, the closed-form solution of \begin{scriptsize}${\bf{L}}_{{l_2}}^c$\end{scriptsize} is}:
\begin{equation}
\scriptsize
\begin{split}
{\left( {{{\bf{L}}_{{l_2}}}} \right)_{m + 1}} = {\left( {{{\bf{L}}_{{l_2}}}} \right)_m} + \rho \left( {{{\left( {{{\bf{C}}_{{l_2}}}} \right)}_{m + 1}} - {{\left( {{{\bf{Z}}_{{l_2}}}} \right)}_{m + 1}}} \right)
\end{split}\label{Update L_{l2}_2}
\end{equation}

For learning bases subproblem, BCD method is used to optimize the objective function, thus the Equation~\ref{LEDL3} with \begin{scriptsize}${\bf{C}}_{{l_2}}^c$\end{scriptsize}, \begin{scriptsize}${\bf{Z}}_{{l_2}}^c$\end{scriptsize} and \begin{scriptsize}${\bf{L}}_{{l_2}}^c$\end{scriptsize} fixed can be rewritten as follows:
\begin{equation}
\scriptsize
\begin{split}
< {{\bf{D}}_{{l_2}}},{\bf{W}},{\bf{A}} >  &= \mathop {\arg \min }\limits_{{{\bf{D}}_{{l_2}}},{\bf{W}},{\bf{A}}} \left\| {{{\bf{C}}_{{l_1}}} - {{\bf{D}}_{{l_2}}}{{\bf{C}}_{{l_2}}}} \right\|_F^2 + 2\varepsilon {\left\| {{{\bf{Z}}_{{l_2}}}} \right\|_{\ell_1}}
\\& + 2{\left( {{{\bf{L}}_{{l_2}}}} \right)^T}\left( {{{\bf{C}}_{{l_2}}} - {{\bf{Z}}_{{l_2}}}} \right) + \rho \left\| {{{\bf{C}}_{{l_2}}} - {{\bf{Z}}_{{l_2}}}} \right\|_F^2
\\&+ \lambda \left\| {{\bf{H}} - {\bf{W}}{{\bf{C}}_{{l_2}}}} \right\|_F^2 + \omega \left\| {{\bf{Q}} - {\bf{A}}{{\bf{C}}_{{l_2}}}} \right\|_F^2
\\s.t.{\kern 1pt} {\kern 1pt} &\left\| {{{\left( {{{\bf{D}}_{{l_2}}}} \right)}_{ \bullet k}}} \right\|_2^2 \le 1,{\kern 1pt} {\kern 1pt} {\kern 1pt} {\kern 1pt} \left\| {{{\bf{W}}_{ \bullet k}}} \right\|_2^2 \le 1,{\kern 1pt} {\kern 1pt} {\kern 1pt} 
\\&\left\| {{{\bf{A}}_{ \bullet k}}} \right\|_2^2 \le 1{\kern 1pt} {\kern 1pt} {\kern 1pt} (k = 1,2, \cdots, K_{{2}})
\end{split}\label{LEDL4}
\end{equation}
To this end, we can solve the closed-form solution with respect to the single column by follows:
\\\\(4) {\bfseries Updating \begin{scriptsize}${\bf{D}}_{{l_2}}$\end{scriptsize} while fixing \begin{scriptsize}${\bf{W}}$\end{scriptsize}, \begin{scriptsize}${\bf{A}}$\end{scriptsize}, \begin{scriptsize}${\bf{C}}_{{l_2}}$\end{scriptsize}, \begin{scriptsize}${\bf{Z}}_{{l_2}}$\end{scriptsize} and \begin{scriptsize}${\bf{L}}_{{l_2}}$\end{scriptsize}, the closed-form solution of \begin{scriptsize}${\bf{D}}_{{l_2}}$\end{scriptsize} is}:
\begin{equation}
\scriptsize
\begin{split}
{\left( {{{\left( {{{\bf{D}}_{{l_2}}}} \right)}_{ \bullet k}}} \right)_{m + 1}} = \frac{{{{{\bf{\dot D}}}_{{l_2}}}}}{{{{\left\| {{{{\bf{\dot D}}}_{{l_2}}}} \right\|}_2}}}
\end{split}\label{Update D_{l2}_2}
\end{equation}
the \begin{scriptsize}${{\bf{\dot D}}_{{l_2}}}$\end{scriptsize} here can be written as:
\begin{equation}
\scriptsize
\begin{split}
{{{\bf{\dot D}}}_{{l_2}}} &= {{\bf{C}}_{{l_1}}}{\left[ {{{\left( {{{\left( {{{\bf{C}}_{{l_2}}}} \right)}_{k \bullet }}} \right)}_{m + 1}}} \right]^T}
- {\left( {{{\left( {{{{\bf{\tilde D}}}_{{l_2}}}} \right)}^k}} \right)_m}{\left( {{{\bf{C}}_{{l_2}}}} \right)_{m + 1}}{\left[ {{{\left( {{{\left( {{{\bf{C}}_{{l_2}}}} \right)}_{k \bullet }}} \right)}_{m + 1}}} \right]^T}
\end{split}\label{Update D_{l2}_3}
\end{equation}
where \begin{scriptsize}$ {\left( {{{{\bf{\tilde D}}}_{{l_2}}}} \right)^k} = \left\{ {\begin{array}{*{20}{c}}
		{{{\left( {{{\bf{D}}_{{l_2}}}} \right)}_{ \bullet p}},{\kern 1pt} {\kern 1pt} {\kern 1pt} {\kern 1pt} {\kern 1pt} {\kern 1pt} {\kern 1pt} {\kern 1pt} p \ne k}\\
		{{\kern 1pt} {\kern 1pt} {\kern 1pt} {\kern 1pt} {\kern 1pt} {\kern 1pt} {\bf{0}},{\kern 1pt} {\kern 1pt} {\kern 1pt} {\kern 1pt} {\kern 1pt} {\kern 1pt} {\kern 1pt} {\kern 1pt} {\kern 1pt} {\kern 1pt} {\kern 1pt} {\kern 1pt} {\kern 1pt} {\kern 1pt} {\kern 1pt} {\kern 1pt} {\kern 1pt} {\kern 1pt} {\kern 1pt} {\kern 1pt} {\kern 1pt} {\kern 1pt} {\kern 1pt} {\kern 1pt} {\kern 1pt} {\kern 1pt} {\kern 1pt} {\kern 1pt} {\kern 1pt} {\kern 1pt} p = k}
		\end{array}} \right.$\end{scriptsize}.
\\\\(5) {\bfseries Updating \begin{scriptsize}${\bf{W}}$\end{scriptsize} while fixing \begin{scriptsize}${\bf{D}}_{{l_2}}$\end{scriptsize}, \begin{scriptsize}${\bf{A}}$\end{scriptsize}, \begin{scriptsize}${\bf{C}}_{{l_2}}$\end{scriptsize}, \begin{scriptsize}${\bf{Z}}_{{l_2}}$ and ${\bf{L}}_{{l_2}}$\end{scriptsize}, the closed-form solution of ${\bf{W}}$ is}:
\begin{equation}
\scriptsize
\begin{split}
\left( {{{\bf{W}}_{ \bullet k}}} \right){{\kern 1pt} _{m + 1}} = \frac{{{\bf{\dot W}}}}{{{{\left\| {{\bf{\dot W}}} \right\|}_2}}}
\end{split}\label{Update W_2}
\end{equation}
the \begin{scriptsize}${{\bf{\dot W}}}$\end{scriptsize} here can be rewritten as:
\begin{equation}
\scriptsize
\begin{split}
{\bf{\dot W}} &= {\bf{H}}{\left[ {{{\left( {{{\left( {{{\bf{C}}_{{l_2}}}} \right)}_{k \bullet }}} \right)}_{m + 1}}} \right]^T} 
- {\left( {{{{\bf{\tilde W}}}^k}} \right)_m}{\left( {{{\bf{C}}_{{l_2}}}} \right)_{m + 1}}{\left[ {{{\left( {{{\left( {{{\bf{C}}_{{l_2}}}} \right)}_{k \bullet }}} \right)}_{m + 1}}} \right]^T}
\end{split}\label{Update W_3}
\end{equation}

where \begin{scriptsize}${{\bf{\tilde W}}^k} = \left\{ {\begin{array}{*{20}{c}}
		{{{\bf{W}}_{ \bullet p}},p \ne k}\\
		{{\bf{0}},{\kern 1pt} {\kern 1pt} {\kern 1pt} {\kern 1pt} {\kern 1pt} {\kern 1pt} {\kern 1pt} {\kern 1pt} {\kern 1pt} {\kern 1pt} {\kern 1pt} p = k}
		\end{array}} \right.$\end{scriptsize};
\\\\(6) {\bfseries Updating ${\bf{A}}$ while fixing \begin{scriptsize}${\bf{D}}_{{l_2}}$\end{scriptsize}, \begin{scriptsize}${\bf{W}}$\end{scriptsize}, \begin{scriptsize}${\bf{C}}_{{l_2}}$\end{scriptsize}, \begin{scriptsize}${\bf{Z}}_{{l_2}}$\end{scriptsize} and \begin{scriptsize}${\bf{L}}_{{l_2}}$\end{scriptsize}, the closed-form solution of \begin{scriptsize}${\bf{A}}$\end{scriptsize} is}:

\begin{equation}
\scriptsize
\begin{split}
\left( {{{\bf{A}}_{ \bullet k}}} \right){{\kern 1pt} _{m + 1}} = \frac{{{\bf{\dot A}}}}{{{{\left\| {{\bf{\dot A}}} \right\|}_2}}}
\end{split}\label{Update A_2}
\end{equation}

The \begin{scriptsize}${\bf{A}}$\end{scriptsize} here can be rewritten as:

\begin{equation}
\scriptsize
\begin{split}
{\bf{\dot A}} &= {\bf{Q}}{\left[ {{{\left( {{{\left( {{{\bf{C}}_{{l_2}}}} \right)}_{k \bullet }}} \right)}_{m + 1}}} \right]^T} 
- {\left( {{{{\bf{\tilde A}}}^k}} \right)_m}{\left( {{{\bf{C}}_{{l_2}}}} \right)_{m + 1}}{\left[ {{{\left( {{{\left( {{{\bf{C}}_{{l_2}}}} \right)}_{k \bullet }}} \right)}_{m + 1}}} \right]^T}
\end{split}\label{Update A_3}
\end{equation}
where \begin{scriptsize}${{\bf{\tilde A}}^k} = \left\{ {\begin{array}{*{20}{c}}
		{{{\bf{A}}_{ \bullet p}},p \ne k}\\
		{{\bf{0}},{\kern 1pt} {\kern 1pt} {\kern 1pt} {\kern 1pt} {\kern 1pt} {\kern 1pt} {\kern 1pt} {\kern 1pt} {\kern 1pt} {\kern 1pt} {\kern 1pt} p = k}
		\end{array}} \right.$\end{scriptsize}.

\subsection{Convergence analysis}

The convergence of CSDL-SRC has been demonstrated in~\cite{liu2016face}.

\begin{figure}[t]
	\begin{center}
		\includegraphics[width=1.0\linewidth]{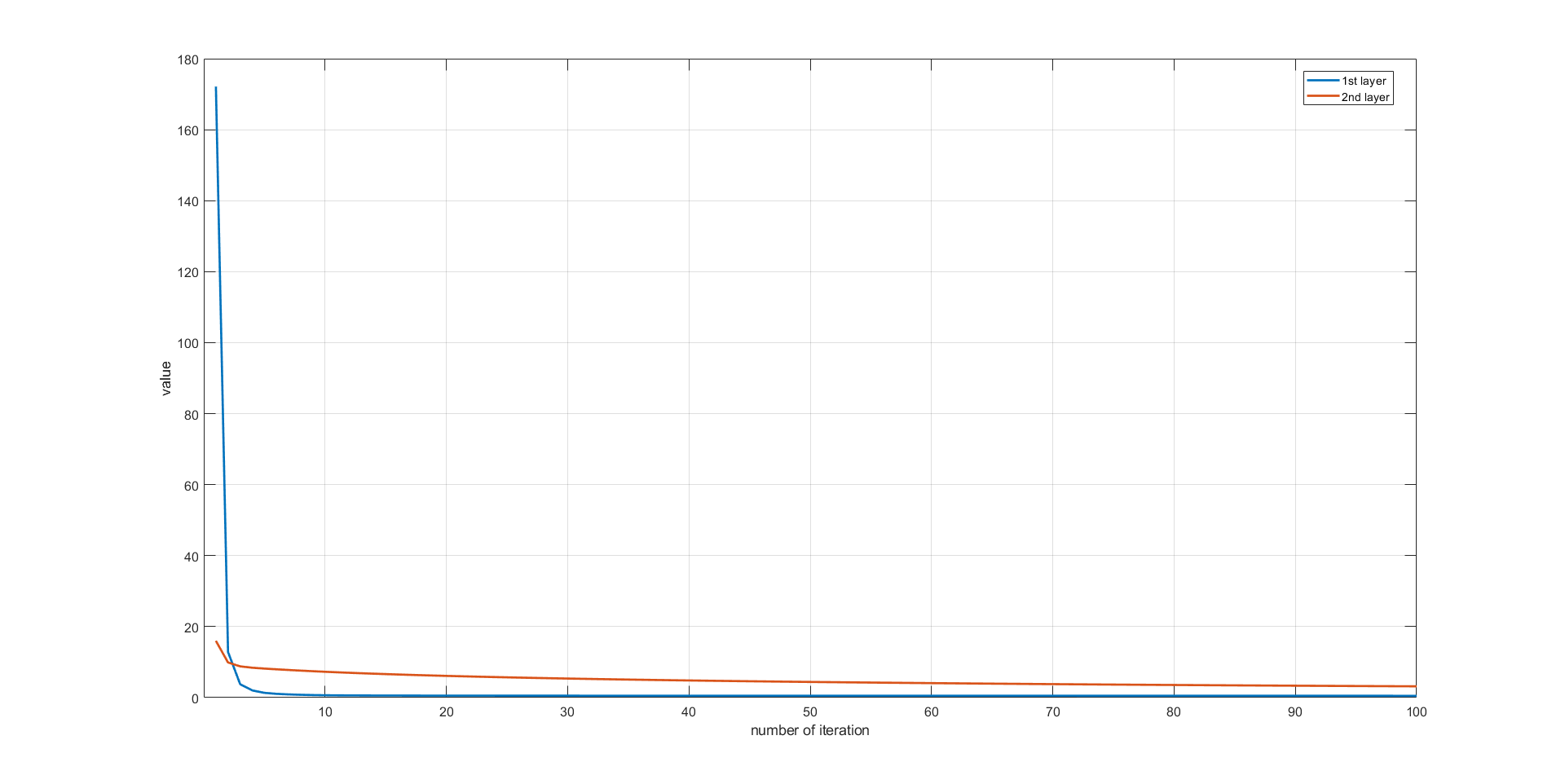}
	\end{center}
	\caption{Convergence curve of CDLF on Extended YaleB dataset}
	\label{fig:Convergence}
\end{figure}

Assume that the result of the objective function after \begin{scriptsize}${m_{th}}$\end{scriptsize} iteration is defined as \begin{scriptsize}$f\left({{{\bf{C}}_m},{{\bf{Z}}_m},{{\bf{L}}_m},{{\bf{B}}_m},{{\bf{W}}_m},{{\bf{A}}_m}} \right)$\end{scriptsize}. 
Since the minimum point is obtained by ADMM and BCD methods, each method will monotonically decrease the corresponding objective function. Considering that the objective function is obviously bounded below and satisfies the Equation (\ref{converge}), it converges. Figure~\ref{fig:Convergence} shows the convergence curve of CDLF by using the Extended YaleB dataset.

\begin{equation}
\scriptsize
\begin{split}
&{\kern 10pt}f\left({{\left( {{{\bf{C}}_{{l_2}}}} \right)_m},{\left( {{{\bf{Z}}_{{l_2}}}} \right)_m},{\left( {{{\bf{L}}_{{l_2}}}} \right)_m},{{\bf{B}}_m},{{\bf{W}}_m},{{\bf{A}}_m}} \right) 
\\&\ge f\left( {{\left( {{{\bf{C}}_{{l_2}}}} \right)_{m + 1}},{\left( {{{\bf{Z}}_{{l_2}}}} \right)_{m + 1}},{\left( {{{\bf{L}}_{{l_2}}}} \right)_{m + 1}},{{\bf{B}}_m},{{\bf{W}}_m},{{\bf{A}}_m}} \right)
\\&\ge f\left( {{\left( {{{\bf{C}}_{{l_2}}}} \right)_{m + 1}},{\left( {{{\bf{Z}}_{{l_2}}}} \right)_{m + 1}},{\left( {{{\bf{L}}_{{l_2}}}} \right)_{m + 1}},{{\bf{B}}_{m + 1}},{{\bf{W}}_{m + 1}},{{\bf{A}}_{m + 1}}} \right)
\end{split}\label{converge}
\end{equation}

\subsection{Overall algorithm}

The overall training procedures of our proposed framework is summarized in Algorithm~\ref{Algorithm1}. Here, \begin{scriptsize}$maxiter$\end{scriptsize} is the maximum number of iterations, \begin{scriptsize}${\bf{1}}\in\mathbb{R}^{K_1\times K_1}$\end{scriptsize} is a square matrix with all elements 1 and \begin{scriptsize}$ \odot $\end{scriptsize} indicates element dot product. \begin{scriptsize}${\bf{Z}}$\end{scriptsize} represents the auxiliary variable, \begin{scriptsize}${\bf{C}}$\end{scriptsize} is another form of \begin{scriptsize}${\bf{S}}$\end{scriptsize} and \begin{scriptsize}${\bf{L}}$\end{scriptsize} is augmented lagrange multiplier. In the algorithm~\ref{Algorithm1}, we first update the parameters of first layer to get the sparse codes \begin{scriptsize}${{\bf{S}}_{{l_1}}}$\end{scriptsize} and dictionary \begin{scriptsize}${{\bf{D}}_{{l_1}}}$\end{scriptsize}. Then \begin{scriptsize}${{\bf{S}}_{{l_1}}}$\end{scriptsize} is treated as one of the inputs of second layer to obtain the corresponding bases \begin{scriptsize}${{\bf{D}}_{{l_2}}}$\end{scriptsize}, \begin{scriptsize}${\bf{W}}$\end{scriptsize}. 

\begin{algorithm}[!t]
	\scriptsize
	\caption{Cascaded Dictionary Learning Framework}\label{Algorithm1}
	\hspace*{0.02in} {\bf Input:} ${{\bf{X}}}\in\mathbb{R}^{d\times N}$, ${\bf{H}}\in\mathbb{R}^{C\times N}$, ${\bf{Q}}\in\mathbb{R}^{K_2\times N}$\\
	\hspace*{0.02in} {\bf Output:}  
	${{\bf{D}}_{{l_1}}}\in\mathbb{R}^{d\times K_{{1}}}$, ${{\bf{D}}_{{l_2}}}\in\mathbb{R}^{d\times K_{{2}}}$,
	${{\bf{W}}}\in\mathbb{R}^{C\times K_{{2}}}$,

	\begin{algorithmic}[1]
		\STATE
		\textbf{Initialize 
		$\left( {{\bf{C}}_{{l_1}}^c} \right)_{0}$, 
		$\left( {{\bf{Z}}_{{l_1}}^c} \right)_{0}$,
		$\left( {{\bf{L}}_{{l_1}}^c} \right)_{0}$
		}
        \STATE
		$m = 0$
		\WHILE 
		{$m \le \max iter$}
		\STATE 
		$m \leftarrow m + 1$
		\STATE 
		\textbf{Update ${{\bf{C}}_{{l_1}}^c}$, ${{\bf{Z}}_{{l_1}}^c}$, ${{\bf{L}}_{{l_1}}^c}$} with \textbf{ADMM}
		\STATE 
		\textbf{Update ${{\bf{D}}_{{l_1}}^c}$} with with \textbf{BCD}		
		\ENDWHILE
		\STATE
		${{\bf{D}}_{{l_1}}} \leftarrow \left[ {{\bf{D}}_{{l_1}}^1,{\bf{D}}_{{l_1}}^2, \cdots ,{\bf{D}}_{{l_1}}^C} \right]$;
		${{\bf{S}}_{{l_1}}} \leftarrow \left[ {{\bf{S}}_{{l_1}}^1,{\bf{S}}_{{l_1}}^2, \cdots ,{\bf{S}}_{{l_1}}^C} \right]$ 	
		\STATE
		\textbf{Initialize
		$\left( {{\bf{C}}_{{l_2}}} \right)_{0}$,
		$\left( {{\bf{Z}}_{{l_2}}} \right)_{0}$,
		$\left( {{\bf{L}}_{{l_2}}} \right)_{0}$,
		$\left( {{\bf{D}}_{{l_2}}} \right)_{0}$,
		$\left( {{\bf{W}}_{{l_2}}} \right)_{0}$,
		$\left( {{\bf{A}}_{{l_2}}} \right)_{0}$
		} 
		\STATE 
		$m = 0$
		\WHILE 
		{$m \le \max iter$}
		\STATE 
		$m \leftarrow m + 1$
		\STATE 
		\textbf{Update ${{\bf{C}}_{{l_2}}}$, ${{\bf{Z}}_{{l_2}}}$, ${{\bf{L}}_{{l_2}}}$} with \textbf{ADMM} 
		\STATE 
		\textbf{Update ${{\bf{D}}_{{l_2}}}, {\bf{W}}, {\bf{A}}$} with \textbf{BCD}\\
		\ENDWHILE
		\RETURN 
		${{\bf{D}}_{{l_1}}}$, 
		${{\bf{D}}_{{l_2}}}$, 
		${\bf{W}}$			 
	\end{algorithmic}
\end{algorithm}

In the testing stage, the constraint terms are based on $\ell_1$-norm sparse constraint. Here, we first exploit the learned dictionary \begin{scriptsize}${\bf{D}}_{{l_1}}$\end{scriptsize} to fit the testing sample \begin{scriptsize}$\bf{y}$\end{scriptsize} and the output is the sparse codes \begin{scriptsize}${{\bf{r}}_{{l_1}}}$\end{scriptsize}. The formulation is shown in Equation~\ref{test_1}. 
\begin{equation}
\scriptsize
	\begin{split}
		{\bf{r}}_{l_1} = \arg {\min _ {\bf{r}_{l_1}}}{\mkern 1mu} \left\{ {\left\| {\bf{y}} - {\bf{D}_{l_1}}{\bf{r}_{l_1}} \right\|_2^2 + 2\alpha {{\left\| {\bf{r}_{l_1}} \right\|}_1}} \right\}
	\end{split}\label{test_1}
\end{equation}
Then the learned dictionary \begin{scriptsize}${\bf{D}}_{{l_2}}$\end{scriptsize} are utilised to fit \begin{scriptsize}${{\bf{r}}_{{l_1}}}$\end{scriptsize} and we can obtain the sparse codes \begin{scriptsize}${{\bf{r}}_{{l_2}}}$\end{scriptsize}. We show the objective function in Equation~\ref{test_2}. 
\begin{equation}
\scriptsize
	\begin{split}
		{\bf{r}}_{l_2} = \arg {\min _ {\bf{r}_{l_2}}}{\mkern 1mu} \left\{ {\left\| {\bf{r}_{l_2}} - {\bf{D}_{l_2}}{\bf{r}_{l_2}} \right\|_2^2 + 2\alpha {{\left\| {\bf{r}_{l_2}} \right\|}_1}} \right\}
	\end{split}\label{test_2}
\end{equation}
At last, we use the trained classfier \begin{scriptsize}${\bf{W}}$\end{scriptsize} to predict the label of \begin{scriptsize}${\bf{y}}$\end{scriptsize} which can be formulated as follows:
\begin{equation}
\scriptsize
	\begin{split}
		id\left( {\bf{y}} \right) = {\max} \left\{ {{\bf{W}}{{\bf{r}}_{{l_2}}}} \right\}
	\end{split}\label{label}
\end{equation}

\section{Experimental results}
\label{Experimental results}

In this section, we evaluate the performance of our approach on several benchmark datasets, including two face datasets (Extented YaleB~\cite{georghiades2001few} dataset, CMU PIE~\cite{sim2002cmu} dataset), two handwritten digit datasets (MNIST~\cite{lecun1998gradient} dataset and USPS~\cite{hull1994database} dataset) and two remote sensing datasets (RSSCB7 dataset~\cite{zou2015deep} and UC Mereced Land Use dataset~\cite{yang2010bag}), then compare it with other famous methods such as SVM~\cite{fan2008liblinear}, SRC~\cite{wright2009robust}, CRC~\cite{zhang2011sparse}, SLRC~\cite{deng2018face}, NRC~\cite{xu2019sparse}, Euler-SRC~\cite{song2018euler}, ADDL~\cite{zhang2018jointly}, LC-PDL~\cite{zhang2019scalable}, FDDL~\cite{yang2011fisher},  LC-KSVD~\cite{jiang2013label}, and CSDL-SRC~\cite{liu2016face}.

For all the experiments, we evaluate our methods by randomly selecting $5$ samples per class for training. 
Besides, to eliminate the randomness, we carry out every experiment $8$ times, and we report the mean of the classification rates. 

For convenience, we fix the dictionary sizes($K_1$ and $K_2$) to twice the number of training samples. 
Moreover, there are four other parameters($\zeta$, $\lambda$, $\omega$, $\varepsilon$) that need to be adjusted to achieve the highest classification rates. The details are shown in the following subsections. In the next subsection, we illustrate the experimental results on the six datasets. Moreover, some discussions are finally listed.

\subsection{Extended YaleB dataset}

\begin{figure}[t]
	\begin{center}
		\includegraphics[width=0.8\linewidth]{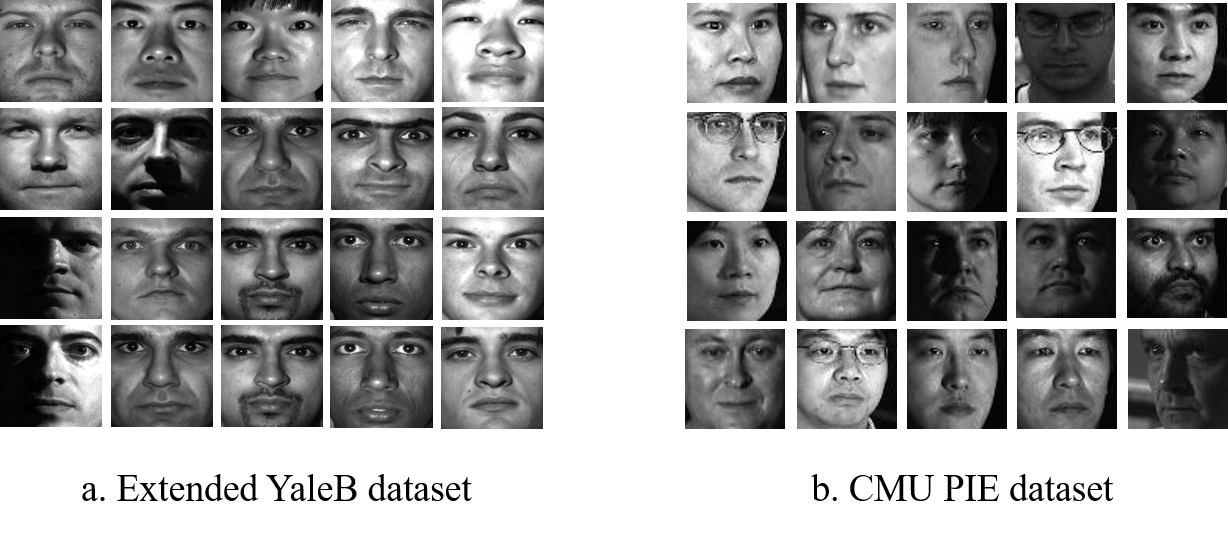}
	\end{center}
	\caption{Examples of the face datasets}
	\label{fig:Face_datasets}
\end{figure}

The Extended YaleB dataset is consists of $2{,}432$ face images from 38 individuals, each having around 64 nearly frontal images under varying illumination conditions. Here, we resize each image to $32 \times 32$ pixels and then pull them into column vectors. After that, we normalize the images to form the raw $\ell_2$ normalized features. Figure~\ref{fig:Face_datasets} $a$ shows some images of the dataset.

In addition, we set $\lambda = {2^{-3}}$, $\omega = {2^{-11}}$ and  $\varepsilon = {2^{ - 8}}$ for LEDL algorithm and set $\zeta = {2^{ - 10}}$, $\lambda = {2^{ - 6}}$, $\omega = {2^{ - 10}}$ and $\varepsilon = {2^{ - 8}}$ in our experiment to achieve highest accuracy for both algorithms, respectively. The experimental results are summarized in Table~\ref{table_face}. From Table~\ref{table_face}, we can see that our proposed LEDL and CDLF algorithms achieve superior performance to other methods. Compared with some conventional algorithms which the DL method is not involved in such as SVM, SRC, CRC, SLRC, NRC, and Euler-SRC, the classification performance is improved by $2.1 \%$ and $2.9\%$ with our proposed LEDL algorithm and CDLF algorithm, respectively. Compared with five classical DL based algorithms, including ADDL, LC-PDL, FDDL, LC-KSVD, and CSDL-SRC, our proposed LEDL algorithm and CDLF algorithm exceeds $1.1\%$ and $1.9\%$, respectively. Additionally, the classification performance of CDLF algorithm exceeds that of LEDL algorithm by $0.8\%$.

\begin{table}
	\scriptsize
	\caption{Classification rates ($\%$) on face datasets}
	\begin{center}
		\begin{tabular}{|c|cc|}
			\hline
			Methods&  Extended YaleB &  CMU PIE\\
			\hline\hline
			SVM~\cite{fan2008liblinear}     & $73.6$ & $71.8$ \\
			SRC~\cite{wright2009robust}     & $79.1$ & $73.7$ \\
			CRC~\cite{zhang2011sparse}      & $79.2$ & $73.3$ \\
			SLRC~\cite{deng2018face}        & $76.7$ & $70.1$ \\
			NRC~\cite{xu2019sparse}         & $76.1$ & $71.0$ \\
			Euler-SRC~\cite{song2018euler}  & $78.5$ & $74.4$ \\			
			ADDL~\cite{zhang2018jointly}    & $77.4$ & $71.1$ \\
			LC-PDL~\cite{zhang2019scalable} & $77.5$ & $70.2$ \\
			FDDL~\cite{yang2011fisher}      & $76.8$ & $70.7$ \\ 
			LC-KSVD~\cite{jiang2013label}   & $73.5$ & $67.1$ \\
			CSDL-SRC~\cite{liu2016face}     & $80.2$ & $77.4$ \\
			Our LEDL                        & $81.3$ & $77.7$ \\
			Our CDLF                        & $\bf82.1$ & $\bf78.7$ \\
			\hline
		\end{tabular}
	\end{center}
	\label{table_face}
\end{table}

\begin{figure*}
	\begin{center}
		\includegraphics[width=1.0\linewidth]{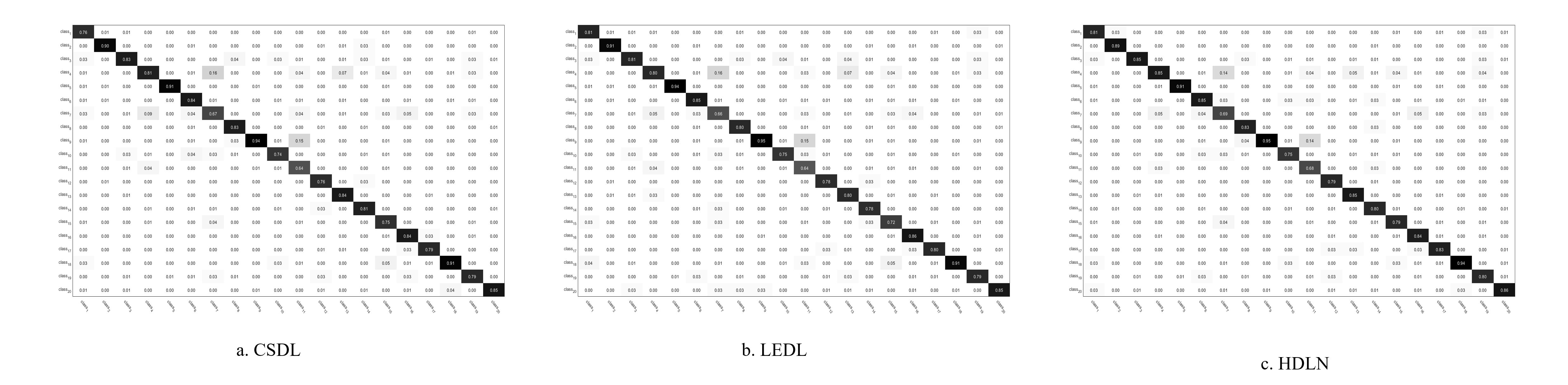}
	\end{center}
	\caption{Confusion matrices on Extended YaleB dataset}
	\label{fig:ConfusionMatrix4ExtendedYaleB}
\end{figure*}

To further illustrate the superiority of our proposed CDLF, we choose the first 20 classes' samples of the Extended YaleB dataset as a sub-dataset to build a confusion matrix. Figure~\ref{fig:ConfusionMatrix4ExtendedYaleB} show the confusion matrices of different methods. As can be seen, our method achieves a higher classification rate in most of the chosen $20$ classes. More specifically, for some classes such as class3, class 4, class 11, class 15, class 17, we get poor classification rates by utilizing CSDL-SRC and LEDL separately. However, there are notable gains while using CDLF. And for some classes (class 1, class 6, class 8, class 9, class 10), which the accuracies have large differences between CSDL-SRC and LEDL, the classification rate of CDLF is similar to the result of the optimal one of CSDL-SRC and LEDL.

\subsection{CMU PIE dataset}

The CMU PIE dataset contains $41{,}368$ images of $68$ individuals with $43$ different illumination conditions. Each human is under $13$ different poses and with $4$ different expressions. Like the Extended YaleB dataset, each face image is cropped to $32 \times 32$ pixels, pulled into column vectors, and normalized to $1$. Several samples from this dataset are listed in Figure~\ref{fig:Face_datasets} $b$.

\begin{figure*}[t]
	\begin{center}
		\includegraphics[width=0.8\linewidth]{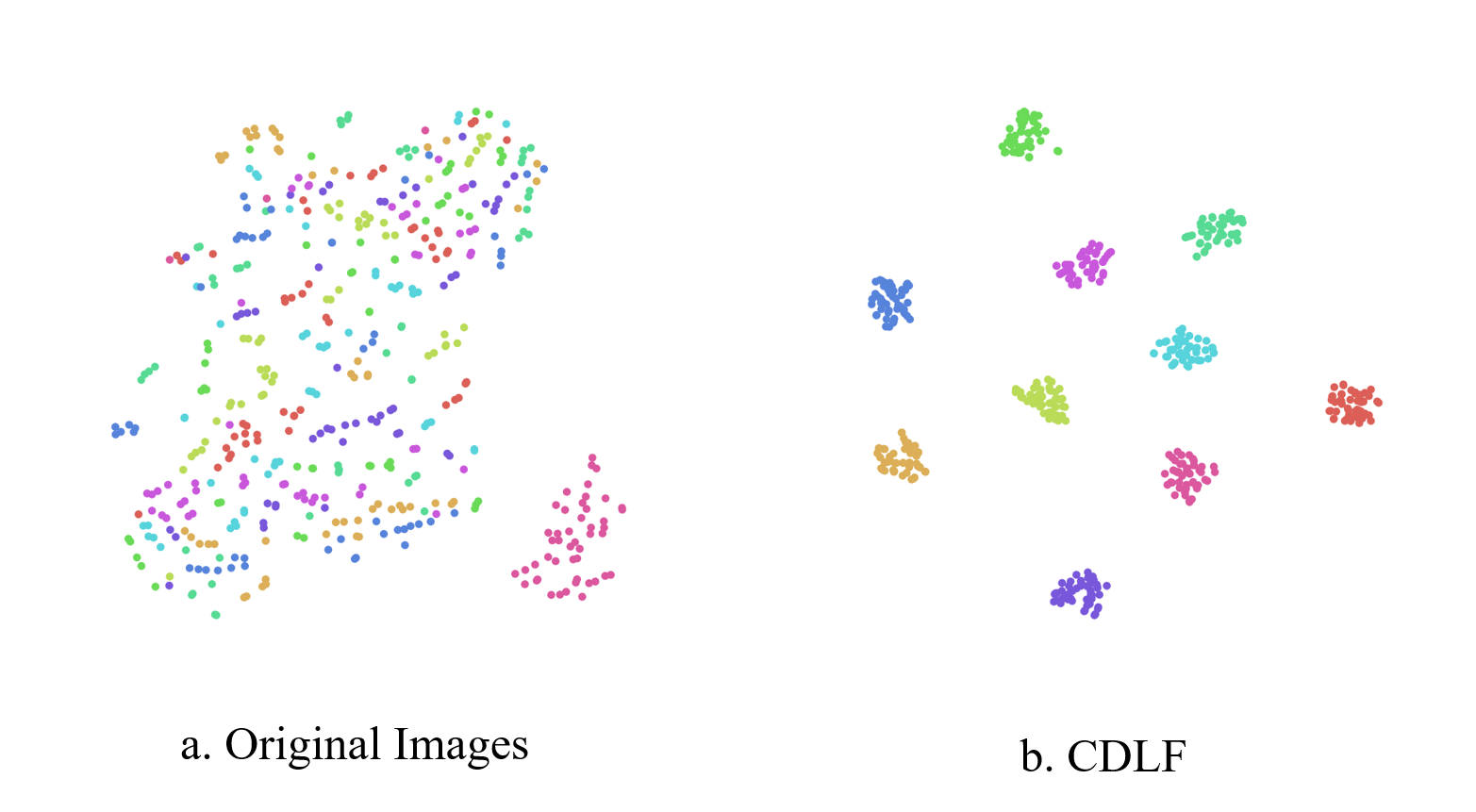}
	\end{center}
	\caption{t-SNE visualization}
	\label{fig:tsne}
\end{figure*}

The results are shown in Table~\ref{table_face}, as can be seen that our methods outperforms all the competing approaches by setting $\lambda = {2^{ - 3}}$, $\omega = {2^{ - 11}}$, $\varepsilon = {2^{ - 8}}$ for LEDL algorithm and $\zeta = {2^{ - 12}}$, $\lambda = {2^{ - 5}}$, $\omega = {2^{ - 11}}$, $\varepsilon = {2^{ - 3}}$ for CDLF algorithm. Specifically, our proposed method achieves an improvement of at least $3.3\%$ and $4.3\%$ over some traditional methods such as SVM, SRC, CRC, SLRC, NRC, and Euler-SRC for LEDL algorithm and CDLF algorithm, respectively. Compared with DL based methods, our proposed LEDL algorithm and CDLF algorithm exceed the other algorithms at least $0.3\%$ and $1.3\%$, respectively. Futhermore, we reduce the dimensionality using t-distributed Stochastic Neighbor Embedding(t-SNE)~\cite{maaten2008visualizing} to show the distribution of the feature extracted from PIE dataset. The results are shown in Figure~\ref{fig:tsne}, it is clear to see the distinction among the categories after our proposed method. 

\begin{figure*}[t]
	\begin{center}
		\includegraphics[width=0.8\linewidth]{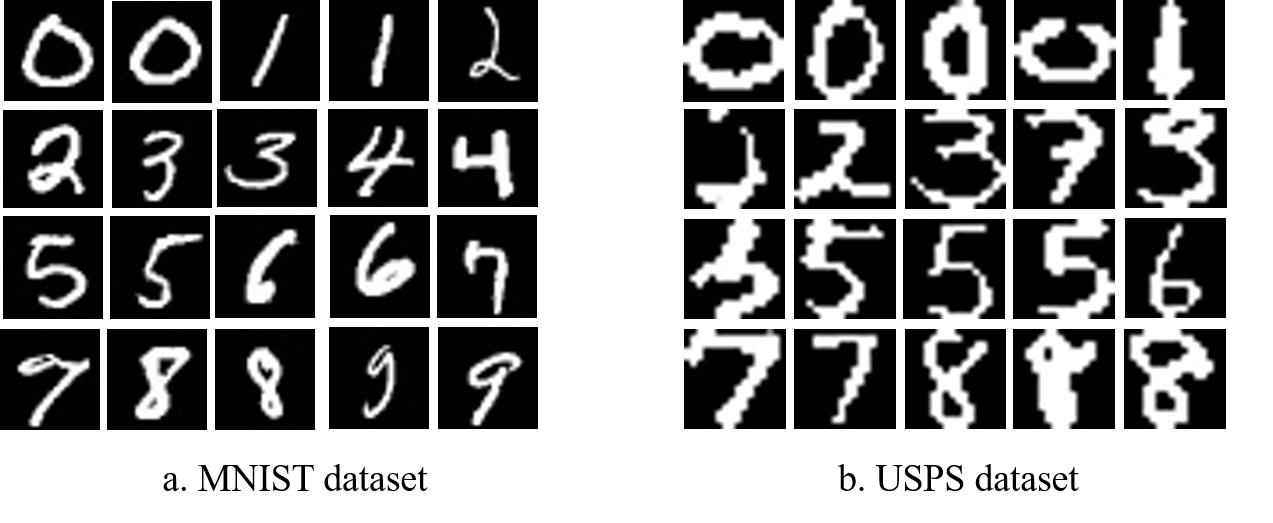}
	\end{center}
	\caption{Examples of the handwritten digit datasets}
	\label{fig:Handwritten_digit_datasets}
\end{figure*}

\begin{table}
	\scriptsize
	\caption{Classification rates ($\%$) on handwritten digit datasets}
	\begin{center}
		\begin{tabular}{|c|cc|}
			\hline
			Methods &  MNIST &  USPS\\
			\hline\hline
			SVM~\cite{fan2008liblinear}     & $65.4$ & $78.8$ \\
			SRC~\cite{wright2009robust}     & $69.8$ & $78.4$ \\
			CRC~\cite{zhang2011sparse}      & $68.3$ & $77.9$ \\
			SLRC~\cite{deng2018face}        & $66.5$ & $76.4$ \\
			NRC~\cite{xu2019sparse}         & $68.5$ & $76.2$ \\
			Euler-SRC~\cite{song2018euler}  & $65.9$ & $76.1$ \\			
			ADDL~\cite{zhang2018jointly}    & $64.9$ & $65.6$ \\
			LC-PDL~\cite{zhang2019scalable} & $60.5$ & $63.2$ \\
			FDDL~\cite{yang2011fisher}      & $62.4$ & $76.2$ \\ 
			LC-KSVD~\cite{jiang2013label}   & $62.1$ & $71.1$ \\
			CSDL-SRC~\cite{liu2016face}     & $69.8$ & $78.8$ \\
			Our LEDL                        & $69.8$ & $81.1$ \\
			Our CDLF                        & $\bf70.3$ & $\bf81.9$ \\
			\hline
		\end{tabular}
	\end{center}
	\label{table_digit}
\end{table}

\subsection{MNIST dataset}

The MNIST dataset includes $70{,}000$ images for digit numbers from $0$ to $9$. Here, we pull the original images in which the size is $28 \times 28$ into column vectors. There are some samples from the dataset are given in Figure~\ref{fig:Handwritten_digit_datasets} $a$. 

In Tabel~\ref{table_digit}, we can see that the classification rates of some conventional methods such as SVM, SRC, CRC, SLRC, NRC, and Euler-SRC can achieve the similar ones of DL based methods (e.g. the classification rates between SRC, CSDL-SRC, and LEDL are similar). However, our proposed CDLF can achieve the highest accuracy by an improvement of at least $0.5\%$ compared with all the methods in Tabel~\ref{table_digit}. The optimal parameter for LEDL algorithm are $\lambda = {2^{ - 8}}$, $\omega = {2^{ - 14}}$, $\varepsilon = {2^{ - 4}}$ and the optimal parameters for CDLF algorithm are $\zeta = {2^{ - 8}}$, $\lambda = {2^{ - 6}}$, $\omega = {2^{ - 6}}$, $\varepsilon = {2^{ - 2}}$.

\subsection{USPS dataset}

The USPS dataset consists of $9{,}298$ handwritten digit images from $0$ to $9$, which come from the U.S. Postal System. For the USPS dataset, the images are resized into $16 \times 16$ and pulled into column vectors.  
Several samples from this dataset are listed in Figure~\ref{fig:Handwritten_digit_datasets} $b$.

The results are showed in Tabel~\ref{table_digit}. For LEDL algorithm, we adjust $\lambda = {2^{ - 4}}$, $\omega = {2^{ - 8}}$, $\varepsilon = {2^{ - 5}}$. For CDLF algorithm, we adjust $\zeta = {2^{ - 11}}$, $\lambda = {2^{ - 10}}$, $\omega = {2^{ - 14}}$, $\varepsilon = {2^{ - 8}}$ to achieve the highest accuracy. Compared with the methods (SVM, SRC, CRC, SLRC, NRC, and Euler-SRC) which the DL is not added into the classifiers, CDLF algorithm achieves an improvement of at least $3.1\%$ and LEDL algorithm achieves an improvement of at least $2.3\%$. Compared with the DL based method, LEDL algorithm achieves an improvement of $3.1\%$.

\subsection{RSSCN7 dataset}

\begin{figure}[t]
	\begin{center}
		\includegraphics[width=0.8\linewidth]{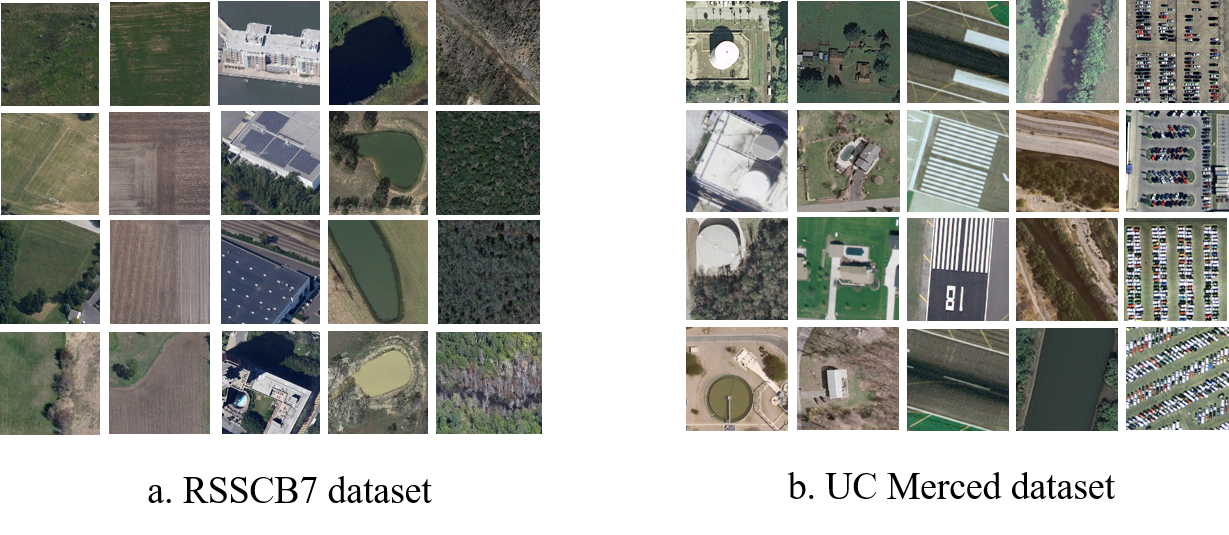}
	\end{center}
	\caption{Examples of the remote sensing datasets}
	\label{fig:Remote_sensing_datasets}
\end{figure}

The RSSCN7 dataset consists of seven different RS scene categories of $2,800$ aerial-scene images in total, which are grassland, forest, farmland, industry, parking lot, residential, river, and lake region. Each class included $400$ images, and all images are of the same size of $400\times400$ pixels. Here, we use the Resnet model~\cite{he2016deep} to extract the features. Specifically, the layer $pool5$ is utilized to extract $2,048$-dimensional vectors for them. Figure~\ref{fig:Remote_sensing_datasets} $a$ shows several samples belongs to this dataset.

\begin{table}
	\scriptsize
	\caption{Classification rates ($\%$) on remote sensing datasets}
	\begin{center}
		\begin{tabular}{|c|cc|}
			\hline
			Methods & RSSCB7  & UC Mereced\\
			\hline\hline
			SVM~\cite{fan2008liblinear}     & $67.5$ & $80.5$ \\
			SRC~\cite{wright2009robust}     & $67.1$ & $80.4$ \\
			CRC~\cite{zhang2011sparse}      & $67.7$ & $80.7$ \\
			SLRC~\cite{deng2018face}        & $66.4$ & $80.9$ \\
			NRC~\cite{xu2019sparse}         & $65.5$ & $79.6$ \\
			Euler-SRC~\cite{song2018euler}  & $67.0$ & $80.6$ \\			
			ADDL~\cite{zhang2018jointly}    & $\bf72.3$ & $\bf83.2$ \\
			LC-PDL~\cite{zhang2019scalable} & $69.7$ & $81.2$ \\
			FDDL~\cite{yang2011fisher}      & $60.0$ & $81.0$ \\ 
			LC-KSVD~\cite{jiang2013label}   & $68.0$ & $79.4$\\
			CSDL-SRC~\cite{liu2016face}     & $66.6$ & $80.5$\\
			Our LEDL                        & $67.9$ & $80.7$\\
			Our CDLF                        & $69.6$ & $81.0$\\
			\hline
		\end{tabular}
	\end{center}
	\label{table_remote}
\end{table}

\begin{table}
	\scriptsize
	\caption{Classification rates ($\%$) on remote sensing datasets}
	\begin{center}
		\begin{tabular}{|c|cccccc|}
			\hline
			Methods & Extended YaleB & CMUPIE & MNIST & USPS & RSSCB7  & UC Mereced\\
			\hline\hline
			SVM~\cite{fan2008liblinear}     & $73.6$ &$71.8$ & $65.4$ & $78.8$ & $67.5$ & $80.5$ \\
			SRC~\cite{wright2009robust}     & $79.1$ &$73.7$ & $69.8$ & $78.4$ & $67.1$ & $80.4$ \\
			CRC~\cite{zhang2011sparse}      & $79.2$ &$73.3$ & $68.3$ & $77.9$ & $67.7$ & $80.7$ \\
			SLRC~\cite{deng2018face}        & $76.7$ &$70.1$ & $65.5$ & $76.4$ & $66.4$ & $80.9$ \\
			NRC~\cite{xu2019sparse}         & $76.1$ &$71.0$ & $68.5$ & $76.2$ & $65.5$ & $79.6$ \\
			Euler-SRC~\cite{song2018euler}  & $78.5$ &$74.4$ & $65.9$ & $76.1$ & $67.0$ & $80.6$ \\			
			ADDL~\cite{zhang2018jointly}    & $77.4$ &$71.1$ & $64.9$ & $65.6$ & $\bf72.3$ & $\bf83.2$ \\
			LC-PDL~\cite{zhang2019scalable} & $77.5$ &$70.2$ & $60.5$ & $63.2$ & $69.7$ & $81.2$ \\
			FDDL~\cite{yang2011fisher}      & $76.8$ &$70.7$ & $62.4$ & $76.2$ & $60.0$ & $81.0$ \\ 
			LC-KSVD~\cite{jiang2013label}   & $73.5$ &$67.1$ & $62.1$ & $71.1$ & $68.0$ & $79.4$\\
			CSDL-SRC~\cite{liu2016face}     & $80.2$ &$77.4$ & $69.8$ & $78.8$ & $66.6$ & $80.5$\\
			 LEDL Shao et al. (2020)                           & $81.3$ &$77.7$ & $69.8$ & $81.1$ & $67.9$ & $80.7$\\
			Our CDLF                        & $\bf82.1$ &$\bf78.7$ & $\bf70.3$ & $\bf81.9$ & $69.6$ & $81.0$\\
			\hline
		\end{tabular}
	\end{center}
	\label{table_remote}
\end{table}

Table~\ref{table_remote} shows the experimental results. It is clearly to see that only three methods (ADDL, LC-PDL, and CSDL) can gain good performance. We will illustrate the analysis why the performance of our methods is lower than that of ADDL in \ref{Analysis and discussion}. The optimal parameters are $\lambda = {2^{ - 8}}$, $\omega = {2^{ - 14}}$, $\varepsilon = {2^{ - 14}}$ for LEDL algorithm. And the optimal parameters are $\zeta = {2^{ - 13}}$, $\lambda = {2^{ - 3}}$, $\omega = {2^{ - 11}}$, $\varepsilon = {2^{ - 11}}$ for CDLF algorithm.

\subsection{UC Mereced Land Use dataset}

The UC Merced Land Use Dataset contains a total of $2,100$ land-use images. The dataset is collected from the United States Geological Survey National Map of 20 U.S. regions. The size of each original image is $256\times 256$ pixels. Here, we also use the Resnet model to obtain 2048-dimensional vectors. Some samples are listed in Figure~\ref{fig:Remote_sensing_datasets} $b$.

Table~\ref{table_remote} shows the classification rates of different methods. Without considering ADDL, it is hard to say that the DL method contributes a lot for image classification in this dataset. Whether LC-KSVD, CSDL-SRC, and LEDL can not get better performance than traditional methods such as SVM, SRC, CRC, SLRC, NRC, and Euler-SRC. Only LC-PDL and our proposed CDLF algorithm can achieve a slight improvement. The comparison with ADDL is summarized in \ref{Analysis and discussion}.

\subsection{Analysis and discussion}
\label{Analysis and discussion}
In this section, we mainly focus on illustrating ablation learning and analysis of the experimental results.

\subsubsection{Ablation learning}
\label{Ablation learning}
To further evaluate the necessity of each term in Equation~\ref{LEDL}, we show the ablation experiments on two face datasets in Figure~\ref{fig: Ablation_experiments}. The results demonstrate that all terms contribute a lot to our approach.

\begin{figure}
	\begin{center}
		\includegraphics[width=0.8\linewidth]{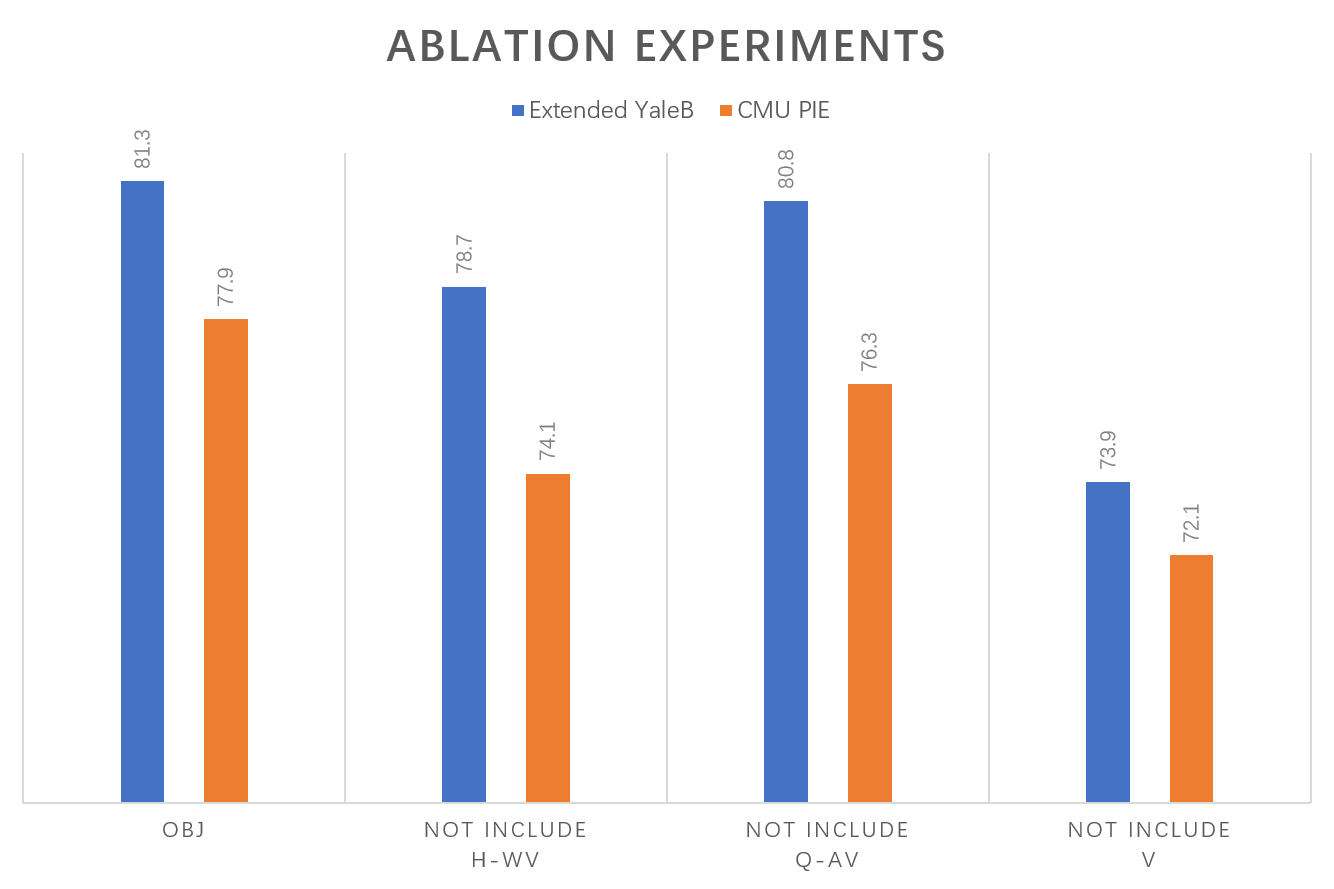}
	\end{center}
	\caption{Ablation experiments in LEDL}
	\label{fig:Ablation_experiments}
\end{figure}

\subsubsection{Analysis of experimental results}
\label{Analysis of experimental results}
We can see that our proposed methods have significant improvement compared with other classical methods except for ADDL on remote sensing datasets. The reason is that ADDL encourages the subdictionaries associated with different classes to be independent while our approaches focus more on global information. Thus, ADDL is more applicable to the datasets, which have significant differences among categories. At the same time, remote sensing datasets usually consist of images that come from different regions. Each class is very different. That is why ADDL can gain better performance than our proposed CDLF on the two specific datasets.

\begin{figure}[h!]
	\begin{center}
	\includegraphics[width = 0.8\linewidth]{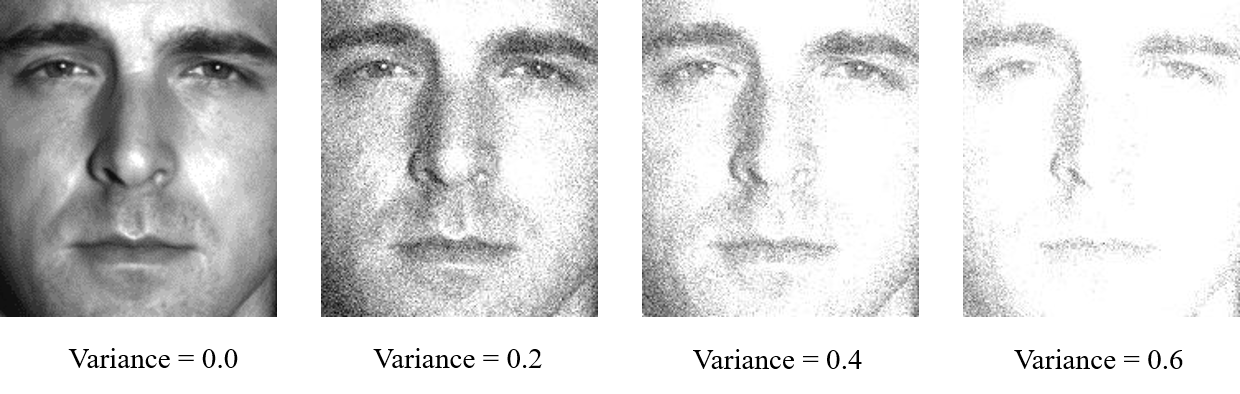}
	\end{center}
	\caption{Examples of the noisy Extended YaleB dataset}
	\label{fig:Gaussian}
\end{figure}

On the other hand, we have demonstrated on various datasets that our proposed CDLF is more general than other methods. However, it is not easy to obtain cleaned datasets in the real world. Besides, the computational complexity of the algorithm is an essential factor to consider in practical applications. To further validate the performance of our proposed methods, we give the following analysis:

1) We add Gaussian noise with different variances ($0.2$, $0.4$, $0.6$) to the Extended YaleB dataset to simulate actual data. Figure~\ref{fig:Gaussian} shows several examples. Table~\ref{tab:classification rates on noisy datasets} shows the experimental results. It is clear to see that our proposed methods are less sensitive to noise than other methods.
    
\begin{table}
	\scriptsize
	\caption{Classification rates ($\%$) on noisy Extended YaleB dataset}
	\begin{center}
		\begin{tabular}{|c|cccc|}
			\hline
			Methods$\backslash$Variance &  0.0 &  0.2 &  0.4 &  0.6 \\
			\hline\hline
			SVM~\cite{fan2008liblinear}     & $73.6$      & $70.5$      & $69.8$      & $67.9$ \\		
			SRC~\cite{wright2009robust}     & $79.1$      & $70.2$      & $60.1$      & $36.7$ \\
			CRC~\cite{zhang2011sparse}      & $79.2$      & $76.2$      & $75.6$      & $74.1$ \\
			SLRC~\cite{deng2018face}        & $76.7$      & $66.4$      & $51.1$      & $19.5$ \\
			NRC~\cite{xu2019sparse}         & $76.1$      & $74.1$      & $70.9$      & $63.3$ \\
			Euler-SRC~\cite{song2018euler}  & $78.5$      & $76.8$      & $76.5$      & $75.5$ \\
			ADDL~\cite{zhang2018jointly}    & $77.4$      & $65.5$      & $51.7$      & $38.6$ \\
			LC-PDL~\cite{zhang2019scalable} & $77.5$      & $65.2$      & $50.9$      & $38.7$ \\
			FDDL~\cite{yang2011fisher}      & $76.8$      & $64.2$      & $46.8$      & $30.0$ \\
			LC-KSVD~\cite{jiang2013label}   & $73.5$      & $74.1$      & $69.4$      & $66.3$ \\
			CSDL-SRC~\cite{liu2016face}     & $80.2$      & $\bf{85.3}$ & $78.1$      & $73.2$ \\
            LEDL                            & $81.3$      & $79.6$      & $79.5$      & $76.1$ \\
            CDLF                            & $\bf{82.1}$ & $80.9$      & $\bf{81.6}$ & $\bf{76.8}$ \\
			\hline
		\end{tabular}
	\end{center}
	\label{tab:classification rates on noisy datasets}
\end{table}
2) We list the computational complexity of some methods in Table~\ref{tab:Complexity}. $N$ is the number of training samples, $D$ represents the dimensions of the sample features, $K$ is the number of atoms in the dictionary, $T$ is the sparsity constraint factor, $K^c$ and $N^c$ represent the $c_{th}$ class of $N$ and $K$, respectively. In our experiments, $K$ is twice the number of $N$ and $K>N>D>T$. From the table, we can say that the complexity of our methods is slightly higher than some traditional methods in some cases, which depends on the size of $K$.
\begin{table}[!t]
  \centering
  \scriptsize
  \caption{Computational Complexity}
  \label{tab:Complexity}
  \begin{tabular}{cc}
    \\[-2mm]
    \hline
    \hline\\[-2mm]
    {\bf \small Methods}   & {\bf\small Complexity}\\
    \hline
    \vspace{1mm}\\[-3mm]
        ${\mathbf{SRC}}$       & $O(N^2D)$\\
    \hline
    \vspace{1mm}\\[-3mm]
        ${\mathbf{CRC}}$       & $O(N^2D)$\\
    \hline
    \vspace{1mm}
    \\[-3mm]    
        ${\mathbf{SLRC}}$      & $O(N^2D)$\\
    \hline
    \vspace{1mm}\\[-3mm]
        ${\mathbf{CSDL-SRC}}$  & $O(K^cN^cD)$\\
    \hline
    \vspace{1mm}\\[-3mm]   
        ${\mathbf{LC-KSVD}}$   & $O(KN^2DT)$\\
    \hline
    \vspace{1mm}\\[-3mm]   
        ${\mathbf{LEDL}}$      & $O(KND)$\\
    \hline
    \vspace{1mm}\\[-3mm]   
        ${\mathbf{CDLF}}$      & $O(KND)$\\
    \hline
    \hline
  \end{tabular}
\end{table}

The analysis above illustrates that our methods are more suitable for practical situations compared to other approaches to some extent.

\section{Conclusion}
\label{Conclusion}

In this paper, we first propose a novel class shared dictionary learning method named label embedded dictionary learning (LEDL). This method introduces the $\ell_1$-norm regularization term to replace the $\ell_0$-norm regularization of LC-KSVD. Then we propose a novel network named hybrid dictionary learning network (CDLF) to combine a class specific dictionary learning method with a class shared dictionary learning method to describe the feature to boost the classification performance. Besides, we adopt the ADMM algorithm to solve the $\ell_1$-norm optimization problem and the BCD algorithm to update the corresponding dictionaries. Finally, extensive experiments on six well-known benchmark datasets have proved the superiority of our proposed LEDL and CDLF methods.


%
\section{Acknowledgment}

This research was funded by the National Natural Science Foundation of China (Grant No. 61402535, No. 61271407), the Natural Science Foundation of Shandong Province, China(Grant No. No. ZR2019MF073, ZR2018MF017, ZR2017MF069), Qingdao Science and Technology Project (No. 17-1-1-8-jch), the Fundamental Research Funds for the Central Universities, China University of Petroleum (East China) (Grant No. 20CX05001A, 17CX02027A), the Open Research Fund from Shandong Provincial Key Laboratory of Computer Network (No. SDKLCN-2018-01), the Major Scientific and Technological Projects of CNPC  (No. ZD2019-183-008), and the Creative Research Team of Young Scholars at Universities in Shandong Province (No.2019KJN019).



\bibliographystyle{elsarticle-harv}
\bibliography{egbib}

\begin{thebibliography}{46}
\expandafter\ifx\csname natexlab\endcsname\relax\def\natexlab#1{#1}\fi
\providecommand{\url}[1]{\texttt{#1}}
\providecommand{\href}[2]{#2}
\providecommand{\path}[1]{#1}
\providecommand{\DOIprefix}{doi:}
\providecommand{\ArXivprefix}{arXiv:}
\providecommand{\URLprefix}{URL: }
\providecommand{\Pubmedprefix}{pmid:}
\providecommand{\doi}[1]{\href{http://dx.doi.org/#1}{\path{#1}}}
\providecommand{\Pubmed}[1]{\href{pmid:#1}{\path{#1}}}
\providecommand{\bibinfo}[2]{#2}
\ifx\xfnm\relax \def\xfnm[#1]{\unskip,\space#1}\fi
\bibitem[{Aharon et~al.(2006)Aharon, Elad, Bruckstein et~al.}]{aharon2006k}
\bibinfo{author}{Aharon, M.}, \bibinfo{author}{Elad, M.},
  \bibinfo{author}{Bruckstein, A.}, et~al., \bibinfo{year}{2006}.
\newblock \bibinfo{title}{K-svd: An algorithm for designing overcomplete
  dictionaries for sparse representation}.
\newblock \bibinfo{journal}{IEEE Transactions on Signal Processing}
  \bibinfo{volume}{54}, \bibinfo{pages}{4311--4322}.
\bibitem[{Akhtar et~al.(2016)Akhtar, Shafait and
  Mian}]{akhtar2016discriminative}
\bibinfo{author}{Akhtar, N.}, \bibinfo{author}{Shafait, F.},
  \bibinfo{author}{Mian, A.}, \bibinfo{year}{2016}.
\newblock \bibinfo{title}{Discriminative bayesian dictionary learning for
  classification}.
\newblock \bibinfo{journal}{IEEE Transactions on Pattern Analysis and Machine
  Intelligence} \bibinfo{volume}{38}, \bibinfo{pages}{2374--2388}.
\bibitem[{Boyd et~al.(2011)Boyd, Parikh, Chu, Peleato and
  Eckstein}]{boyd2011distributed}
\bibinfo{author}{Boyd, S.}, \bibinfo{author}{Parikh, N.}, \bibinfo{author}{Chu,
  E.}, \bibinfo{author}{Peleato, B.}, \bibinfo{author}{Eckstein, J.},
  \bibinfo{year}{2011}.
\newblock \bibinfo{title}{Distributed optimization and statistical learning via
  the alternating direction method of multipliers}.
\newblock \bibinfo{journal}{Foundations and Trends{\textregistered} in Machine
  learning} \bibinfo{volume}{3}, \bibinfo{pages}{1--122}.
\bibitem[{Chan et~al.(2015)Chan, Jia, Gao, Lu, Zeng and Ma}]{chan2015pcanet}
\bibinfo{author}{Chan, T.H.}, \bibinfo{author}{Jia, K.}, \bibinfo{author}{Gao,
  S.}, \bibinfo{author}{Lu, J.}, \bibinfo{author}{Zeng, Z.},
  \bibinfo{author}{Ma, Y.}, \bibinfo{year}{2015}.
\newblock \bibinfo{title}{Pcanet: A simple deep learning baseline for image
  classification?}
\newblock \bibinfo{journal}{IEEE Transactions on Image Processing}
  \bibinfo{volume}{24}, \bibinfo{pages}{5017--5032}.
\bibitem[{Deng et~al.(2018)Deng, Hu and Guo}]{deng2018face}
\bibinfo{author}{Deng, W.}, \bibinfo{author}{Hu, J.}, \bibinfo{author}{Guo,
  J.}, \bibinfo{year}{2018}.
\newblock \bibinfo{title}{Face recognition via collaborative representation:
  Its discriminant nature and superposed representation}.
\newblock \bibinfo{journal}{IEEE Transactions on Pattern Analysis and Machine
  Intelligence} \bibinfo{volume}{40}, \bibinfo{pages}{2513--2521}.
\bibitem[{Fan et~al.(2008)Fan, Chang, Hsieh, Wang and Lin}]{fan2008liblinear}
\bibinfo{author}{Fan, R.E.}, \bibinfo{author}{Chang, K.W.},
  \bibinfo{author}{Hsieh, C.J.}, \bibinfo{author}{Wang, X.R.},
  \bibinfo{author}{Lin, C.J.}, \bibinfo{year}{2008}.
\newblock \bibinfo{title}{Liblinear: A library for large linear
  classification}.
\newblock \bibinfo{journal}{Journal of Machine Learning Research}
  \bibinfo{volume}{9}, \bibinfo{pages}{1871--1874}.
\bibitem[{Gao et~al.(2018)Gao, Hu and Ye}]{gao2018self}
\bibinfo{author}{Gao, D.}, \bibinfo{author}{Hu, Z.}, \bibinfo{author}{Ye, R.},
  \bibinfo{year}{2018}.
\newblock \bibinfo{title}{Self-dictionary regression for hyperspectral image
  super-resolution}.
\newblock \bibinfo{journal}{Remote Sensing} \bibinfo{volume}{10},
  \bibinfo{pages}{1574--1596}.
\bibitem[{Georghiades et~al.(2001)Georghiades, Belhumeur and
  Kriegman}]{georghiades2001few}
\bibinfo{author}{Georghiades, A.S.}, \bibinfo{author}{Belhumeur, P.N.},
  \bibinfo{author}{Kriegman, D.J.}, \bibinfo{year}{2001}.
\newblock \bibinfo{title}{From few to many: Illumination cone models for face
  recognition under variable lighting and pose}.
\newblock \bibinfo{journal}{IEEE Transactions on Pattern Analysis and Machine
  Intelligence} \bibinfo{volume}{23}, \bibinfo{pages}{643--660}.
\bibitem[{Gu et~al.(2014)Gu, Zhang, Zuo and Feng}]{gu2014projective}
\bibinfo{author}{Gu, S.}, \bibinfo{author}{Zhang, L.}, \bibinfo{author}{Zuo,
  W.}, \bibinfo{author}{Feng, X.}, \bibinfo{year}{2014}.
\newblock \bibinfo{title}{Projective dictionary pair learning for pattern
  classification}, in: \bibinfo{booktitle}{Advances in neural information
  processing systems}, pp. \bibinfo{pages}{793--801}.
\bibitem[{He et~al.(2016)He, Zhang, Ren and Sun}]{he2016deep}
\bibinfo{author}{He, K.}, \bibinfo{author}{Zhang, X.}, \bibinfo{author}{Ren,
  S.}, \bibinfo{author}{Sun, J.}, \bibinfo{year}{2016}.
\newblock \bibinfo{title}{Deep residual learning for image recognition}, in:
  \bibinfo{booktitle}{IEEE conference on Computer Vision and Pattern
  Recognition (CVPR)}, \bibinfo{organization}{IEEE}. pp.
  \bibinfo{pages}{770--778}.
\bibitem[{Hull(1994)}]{hull1994database}
\bibinfo{author}{Hull, J.J.}, \bibinfo{year}{1994}.
\newblock \bibinfo{title}{A database for handwritten text recognition
  research}.
\newblock \bibinfo{journal}{IEEE Transactions on Pattern Analysis and Machine
  Intelligence} \bibinfo{volume}{16}, \bibinfo{pages}{550--554}.
\bibitem[{Jiang et~al.(2019)Jiang, Wang, Yi, Wang, Lu and
  Jiang}]{jiang2019edge}
\bibinfo{author}{Jiang, K.}, \bibinfo{author}{Wang, Z.}, \bibinfo{author}{Yi,
  P.}, \bibinfo{author}{Wang, G.}, \bibinfo{author}{Lu, T.},
  \bibinfo{author}{Jiang, J.}, \bibinfo{year}{2019}.
\newblock \bibinfo{title}{Edge-enhanced gan for remote sensing image
  superresolution}.
\newblock \bibinfo{journal}{IEEE Transactions on Geoscience and Remote Sensing
  (TGRS)} \bibinfo{volume}{57}, \bibinfo{pages}{5799--5812}.
\bibitem[{Jiang et~al.(2017)Jiang, Zhang, Qin, Zhao, Li and
  Yan}]{jiang2017robust}
\bibinfo{author}{Jiang, W.}, \bibinfo{author}{Zhang, Z.}, \bibinfo{author}{Qin,
  J.}, \bibinfo{author}{Zhao, M.}, \bibinfo{author}{Li, F.},
  \bibinfo{author}{Yan, S.}, \bibinfo{year}{2017}.
\newblock \bibinfo{title}{Robust projective dictionary learning by joint label
  embedding and classification}, in: \bibinfo{booktitle}{2017 IEEE
  International Conference on Data Mining Workshops (ICDMW)},
  \bibinfo{organization}{IEEE}. pp. \bibinfo{pages}{510--517}.
\bibitem[{Jiang et~al.(2013)Jiang, Lin and Davis}]{jiang2013label}
\bibinfo{author}{Jiang, Z.}, \bibinfo{author}{Lin, Z.}, \bibinfo{author}{Davis,
  L.S.}, \bibinfo{year}{2013}.
\newblock \bibinfo{title}{Label consistent k-svd: Learning a discriminative
  dictionary for recognition}.
\newblock \bibinfo{journal}{IEEE Transactions on Pattern Analysis and Machine
  Intelligence} \bibinfo{volume}{35}, \bibinfo{pages}{2651--2664}.
\bibitem[{LeCun et~al.(1998)LeCun, Bottou, Bengio, Haffner
  et~al.}]{lecun1998gradient}
\bibinfo{author}{LeCun, Y.}, \bibinfo{author}{Bottou, L.},
  \bibinfo{author}{Bengio, Y.}, \bibinfo{author}{Haffner, P.}, et~al.,
  \bibinfo{year}{1998}.
\newblock \bibinfo{title}{Gradient-based learning applied to document
  recognition}.
\newblock \bibinfo{journal}{Proceedings of the IEEE} \bibinfo{volume}{86},
  \bibinfo{pages}{2278--2324}.
\bibitem[{Li et~al.(2018)Li, He, Tao, Tang and Wang}]{li2018joint}
\bibinfo{author}{Li, H.}, \bibinfo{author}{He, X.}, \bibinfo{author}{Tao, D.},
  \bibinfo{author}{Tang, Y.}, \bibinfo{author}{Wang, R.}, \bibinfo{year}{2018}.
\newblock \bibinfo{title}{Joint medical image fusion, denoising and enhancement
  via discriminative low-rank sparse dictionaries learning}.
\newblock \bibinfo{journal}{Pattern Recognition} \bibinfo{volume}{79},
  \bibinfo{pages}{130--146}.
\bibitem[{Li et~al.(2012)Li, Fang and Yin}]{li2012efficient}
\bibinfo{author}{Li, S.}, \bibinfo{author}{Fang, L.}, \bibinfo{author}{Yin,
  H.}, \bibinfo{year}{2012}.
\newblock \bibinfo{title}{An efficient dictionary learning algorithm and its
  application to 3-d medical image denoising}.
\newblock \bibinfo{journal}{IEEE Transactions on Biomedical Engineering}
  \bibinfo{volume}{59}, \bibinfo{pages}{417--427}.
\bibitem[{Li et~al.(2019)Li, Zhang, Qin, Zhang and Shao}]{li2019discriminative}
\bibinfo{author}{Li, Z.}, \bibinfo{author}{Zhang, Z.}, \bibinfo{author}{Qin,
  J.}, \bibinfo{author}{Zhang, Z.}, \bibinfo{author}{Shao, L.},
  \bibinfo{year}{2019}.
\newblock \bibinfo{title}{Discriminative fisher embedding dictionary learning
  algorithm for object recognition}.
\newblock \bibinfo{journal}{IEEE Transactions on Neural Networks and Learning
  Systems} .
\bibitem[{Lin et~al.(2018)Lin, Yang, Yang, Shen and Xie}]{lin2018robust}
\bibinfo{author}{Lin, G.}, \bibinfo{author}{Yang, M.}, \bibinfo{author}{Yang,
  J.}, \bibinfo{author}{Shen, L.}, \bibinfo{author}{Xie, W.},
  \bibinfo{year}{2018}.
\newblock \bibinfo{title}{Robust, discriminative and comprehensive dictionary
  learning for face recognition}.
\newblock \bibinfo{journal}{Pattern Recognition} \bibinfo{volume}{81},
  \bibinfo{pages}{341--356}.
\bibitem[{Liu et~al.(2019)Liu, Jing, Li, Yu, Gittens and
  Mahoney}]{liu2019group}
\bibinfo{author}{Liu, B.}, \bibinfo{author}{Jing, L.}, \bibinfo{author}{Li,
  J.}, \bibinfo{author}{Yu, J.}, \bibinfo{author}{Gittens, A.},
  \bibinfo{author}{Mahoney, M.W.}, \bibinfo{year}{2019}.
\newblock \bibinfo{title}{Group collaborative representation for image set
  classification}.
\newblock \bibinfo{journal}{International Journal of Computer Vision}
  \bibinfo{volume}{127}, \bibinfo{pages}{181--206}.
\bibitem[{Liu et~al.(2017)Liu, Gui, Wang, Wang, Shen, Li and
  Wang}]{liu2017class}
\bibinfo{author}{Liu, B.D.}, \bibinfo{author}{Gui, L.}, \bibinfo{author}{Wang,
  Y.}, \bibinfo{author}{Wang, Y.X.}, \bibinfo{author}{Shen, B.},
  \bibinfo{author}{Li, X.}, \bibinfo{author}{Wang, Y.J.}, \bibinfo{year}{2017}.
\newblock \bibinfo{title}{Class specific centralized dictionary learning for
  face recognition}.
\newblock \bibinfo{journal}{Multimedia Tools and Applications}
  \bibinfo{volume}{76}, \bibinfo{pages}{4159--4177}.
\bibitem[{Liu et~al.(2016)Liu, Shen, Gui, Wang, Li, Yan and Wang}]{liu2016face}
\bibinfo{author}{Liu, B.D.}, \bibinfo{author}{Shen, B.}, \bibinfo{author}{Gui,
  L.}, \bibinfo{author}{Wang, Y.X.}, \bibinfo{author}{Li, X.},
  \bibinfo{author}{Yan, F.}, \bibinfo{author}{Wang, Y.J.},
  \bibinfo{year}{2016}.
\newblock \bibinfo{title}{Face recognition using class specific dictionary
  learning for sparse representation and collaborative representation}.
\newblock \bibinfo{journal}{Neurocomputing} \bibinfo{volume}{204},
  \bibinfo{pages}{198--210}.
\bibitem[{Liu et~al.(2014)Liu, Wang, Shen, Zhang and Wang}]{liu2014blockwise}
\bibinfo{author}{Liu, B.D.}, \bibinfo{author}{Wang, Y.X.},
  \bibinfo{author}{Shen, B.}, \bibinfo{author}{Zhang, Y.J.},
  \bibinfo{author}{Wang, Y.J.}, \bibinfo{year}{2014}.
\newblock \bibinfo{title}{Blockwise coordinate descent schemes for sparse
  representation}, in: \bibinfo{booktitle}{IEEE International conference on
  Acoustics, Speech and Signal Processing (ICASSP)},
  \bibinfo{organization}{IEEE}. pp. \bibinfo{pages}{5267--5271}.
\bibitem[{Maaten and Hinton(2008)}]{maaten2008visualizing}
\bibinfo{author}{Maaten, L.v.d.}, \bibinfo{author}{Hinton, G.},
  \bibinfo{year}{2008}.
\newblock \bibinfo{title}{Visualizing data using t-sne}.
\newblock \bibinfo{journal}{Journal of machine learning research}
  \bibinfo{volume}{9}, \bibinfo{pages}{2579--2605}.
\bibitem[{Mallat and Zhang(1993)}]{mallat1993matching}
\bibinfo{author}{Mallat, S.G.}, \bibinfo{author}{Zhang, Z.},
  \bibinfo{year}{1993}.
\newblock \bibinfo{title}{Matching pursuit with time-frequency dictionaries}.
\newblock \bibinfo{journal}{IEEE Transactions on Signal Processing}
  \bibinfo{volume}{41}, \bibinfo{pages}{3397--3415}.
\bibitem[{Olshausen and Field(1996)}]{olshausen1996emergence}
\bibinfo{author}{Olshausen, B.A.}, \bibinfo{author}{Field, D.J.},
  \bibinfo{year}{1996}.
\newblock \bibinfo{title}{Emergence of simple-cell receptive field properties
  by learning a sparse code for natural images}.
\newblock \bibinfo{journal}{Nature} \bibinfo{volume}{381},
  \bibinfo{pages}{607--609}.
\bibitem[{Olshausen and Field(1997)}]{olshausen1997sparse}
\bibinfo{author}{Olshausen, B.A.}, \bibinfo{author}{Field, D.J.},
  \bibinfo{year}{1997}.
\newblock \bibinfo{title}{Sparse coding with an overcomplete basis set: a
  strategy employed by v1?}
\newblock \bibinfo{journal}{Vision Research} \bibinfo{volume}{37},
  \bibinfo{pages}{3311--3325}.
\bibitem[{Ren et~al.(2020)Ren, Zhang, Li, Wang, Liu, Yan and
  Wang}]{ren2020learning}
\bibinfo{author}{Ren, J.}, \bibinfo{author}{Zhang, Z.}, \bibinfo{author}{Li,
  S.}, \bibinfo{author}{Wang, Y.}, \bibinfo{author}{Liu, G.},
  \bibinfo{author}{Yan, S.}, \bibinfo{author}{Wang, M.}, \bibinfo{year}{2020}.
\newblock \bibinfo{title}{Learning hybrid representation by robust dictionary
  learning in factorized compressed space}.
\newblock \bibinfo{journal}{IEEE Transactions on Image Processing} .
\bibitem[{Sim et~al.(2002)Sim, Baker and Bsat}]{sim2002cmu}
\bibinfo{author}{Sim, T.}, \bibinfo{author}{Baker, S.}, \bibinfo{author}{Bsat,
  M.}, \bibinfo{year}{2002}.
\newblock \bibinfo{title}{The cmu pose, illumination, and expression (pie)
  database}, in: \bibinfo{booktitle}{IEEE International conference on Automatic
  Face and Gesture Recognition (FG)}, \bibinfo{organization}{IEEE}. pp.
  \bibinfo{pages}{53--58}.
\bibitem[{Song et~al.(2018)Song, Liu, Gao, Gao, Nie and Cui}]{song2018euler}
\bibinfo{author}{Song, Y.}, \bibinfo{author}{Liu, Y.}, \bibinfo{author}{Gao,
  Q.}, \bibinfo{author}{Gao, X.}, \bibinfo{author}{Nie, F.},
  \bibinfo{author}{Cui, R.}, \bibinfo{year}{2018}.
\newblock \bibinfo{title}{Euler label consistent k-svd for image classification
  and action recognition}.
\newblock \bibinfo{journal}{Neurocomputing} \bibinfo{volume}{310},
  \bibinfo{pages}{277--286}.
\bibitem[{Sun et~al.(2020)Sun, Zhang, Jiang, Zhang, Zhang, Yan and
  Wang}]{sun2020discriminative}
\bibinfo{author}{Sun, Y.}, \bibinfo{author}{Zhang, Z.}, \bibinfo{author}{Jiang,
  W.}, \bibinfo{author}{Zhang, Z.}, \bibinfo{author}{Zhang, L.},
  \bibinfo{author}{Yan, S.}, \bibinfo{author}{Wang, M.}, \bibinfo{year}{2020}.
\newblock \bibinfo{title}{Discriminative local sparse representation by robust
  adaptive dictionary pair learning}.
\newblock \bibinfo{journal}{IEEE Transactions on Neural Networks and Learning
  Systems} .
\bibitem[{Tropp and Gilbert(2007)}]{tropp2007signal}
\bibinfo{author}{Tropp, J.A.}, \bibinfo{author}{Gilbert, A.C.},
  \bibinfo{year}{2007}.
\newblock \bibinfo{title}{Signal recovery from random measurements via
  orthogonal matching pursuit}.
\newblock \bibinfo{journal}{IEEE Transactions on Information Theory}
  \bibinfo{volume}{53}, \bibinfo{pages}{4655--4666}.
\bibitem[{Wright et~al.(2009)Wright, Yang, Ganesh, Sastry and
  Ma}]{wright2009robust}
\bibinfo{author}{Wright, J.}, \bibinfo{author}{Yang, A.Y.},
  \bibinfo{author}{Ganesh, A.}, \bibinfo{author}{Sastry, S.S.},
  \bibinfo{author}{Ma, Y.}, \bibinfo{year}{2009}.
\newblock \bibinfo{title}{Robust face recognition via sparse representation}.
\newblock \bibinfo{journal}{IEEE Transactions on Pattern Analysis and Machine
  Intelligence} \bibinfo{volume}{31}, \bibinfo{pages}{210--227}.
\bibitem[{Xu et~al.(2019)Xu, An, Zhang and Zhang}]{xu2019sparse}
\bibinfo{author}{Xu, J.}, \bibinfo{author}{An, W.}, \bibinfo{author}{Zhang,
  L.}, \bibinfo{author}{Zhang, D.}, \bibinfo{year}{2019}.
\newblock \bibinfo{title}{Sparse, collaborative, or nonnegative representation:
  Which helps pattern classification?}
\newblock \bibinfo{journal}{Pattern Recognition} \bibinfo{volume}{88},
  \bibinfo{pages}{679--688}.
\bibitem[{Yang et~al.(2009)Yang, Yu, Gong and Huang}]{yang2009linear}
\bibinfo{author}{Yang, J.}, \bibinfo{author}{Yu, K.}, \bibinfo{author}{Gong,
  Y.}, \bibinfo{author}{Huang, T.}, \bibinfo{year}{2009}.
\newblock \bibinfo{title}{Linear spatial pyramid matching using sparse coding
  for image classification}, in: \bibinfo{booktitle}{IEEE conference on
  Computer Vision and Pattern Recognition (CVPR)},
  \bibinfo{organization}{IEEE}. pp. \bibinfo{pages}{1794--1801}.
\bibitem[{Yang et~al.(2011)Yang, Zhang, Feng and Zhang}]{yang2011fisher}
\bibinfo{author}{Yang, M.}, \bibinfo{author}{Zhang, L.}, \bibinfo{author}{Feng,
  X.}, \bibinfo{author}{Zhang, D.}, \bibinfo{year}{2011}.
\newblock \bibinfo{title}{Fisher discrimination dictionary learning for sparse
  representation}, in: \bibinfo{booktitle}{International Conference on Computer
  Vision}, \bibinfo{organization}{IEEE}. pp. \bibinfo{pages}{543--550}.
\bibitem[{Yang and Newsam(2010)}]{yang2010bag}
\bibinfo{author}{Yang, Y.}, \bibinfo{author}{Newsam, S.}, \bibinfo{year}{2010}.
\newblock \bibinfo{title}{Bag-of-visual-words and spatial extensions for
  land-use classification}, in: \bibinfo{booktitle}{ACM International
  conference on Advances in Geographic Information Systems (GIS)},
  \bibinfo{organization}{ACM}. pp. \bibinfo{pages}{270--279}.
\bibitem[{Zhang et~al.(2011)Zhang, Yang and Feng}]{zhang2011sparse}
\bibinfo{author}{Zhang, L.}, \bibinfo{author}{Yang, M.}, \bibinfo{author}{Feng,
  X.}, \bibinfo{year}{2011}.
\newblock \bibinfo{title}{Sparse representation or collaborative
  representation: Which helps face recognition?}, in: \bibinfo{booktitle}{IEEE
  International conference on Computer Vision (ICCV)},
  \bibinfo{organization}{IEEE}. pp. \bibinfo{pages}{471--478}.
\bibitem[{Zhang and Li(2010)}]{zhang2010discriminative}
\bibinfo{author}{Zhang, Q.}, \bibinfo{author}{Li, B.}, \bibinfo{year}{2010}.
\newblock \bibinfo{title}{Discriminative k-svd for dictionary learning in face
  recognition}, in: \bibinfo{booktitle}{IEEE conference on Computer Vision and
  Pattern Recognition (CVPR)}, \bibinfo{organization}{IEEE}. pp.
  \bibinfo{pages}{2691--2698}.
\bibitem[{Zhang et~al.(2018)Zhang, Jiang, Qin, Zhang, Li, Zhang and
  Yan}]{zhang2018jointly}
\bibinfo{author}{Zhang, Z.}, \bibinfo{author}{Jiang, W.}, \bibinfo{author}{Qin,
  J.}, \bibinfo{author}{Zhang, L.}, \bibinfo{author}{Li, F.},
  \bibinfo{author}{Zhang, M.}, \bibinfo{author}{Yan, S.}, \bibinfo{year}{2018}.
\newblock \bibinfo{title}{Jointly learning structured analysis discriminative
  dictionary and analysis multiclass classifier}.
\newblock \bibinfo{journal}{IEEE Transactions on Neural Networks and Learning
  Systems} \bibinfo{volume}{29}, \bibinfo{pages}{3798--3814}.
\bibitem[{Zhang et~al.(2019a)Zhang, Jiang, Zhang, Li, Liu and
  Qin}]{zhang2019scalable}
\bibinfo{author}{Zhang, Z.}, \bibinfo{author}{Jiang, W.},
  \bibinfo{author}{Zhang, Z.}, \bibinfo{author}{Li, S.}, \bibinfo{author}{Liu,
  G.}, \bibinfo{author}{Qin, J.}, \bibinfo{year}{2019}a.
\newblock \bibinfo{title}{Scalable block-diagonal locality-constrained
  projective dictionary learning}, in: \bibinfo{booktitle}{International Joint
  Conference on Artificial Intelligence (IJCAI)}, pp.
  \bibinfo{pages}{4376--4382}.
\bibitem[{Zhang et~al.(2016)Zhang, Li, Chow, Zhang and Yan}]{zhang2016sparse}
\bibinfo{author}{Zhang, Z.}, \bibinfo{author}{Li, F.}, \bibinfo{author}{Chow,
  T.W.}, \bibinfo{author}{Zhang, L.}, \bibinfo{author}{Yan, S.},
  \bibinfo{year}{2016}.
\newblock \bibinfo{title}{Sparse codes auto-extractor for classification: A
  joint embedding and dictionary learning framework for representation}.
\newblock \bibinfo{journal}{IEEE Transactions on Signal Processing}
  \bibinfo{volume}{64}, \bibinfo{pages}{3790--3805}.
\bibitem[{Zhang et~al.(2019b)Zhang, Ren, Jiang, Zhang, Hong, Yan and
  Wang}]{zhang2019joint}
\bibinfo{author}{Zhang, Z.}, \bibinfo{author}{Ren, J.}, \bibinfo{author}{Jiang,
  W.}, \bibinfo{author}{Zhang, Z.}, \bibinfo{author}{Hong, R.},
  \bibinfo{author}{Yan, S.}, \bibinfo{author}{Wang, M.}, \bibinfo{year}{2019}b.
\newblock \bibinfo{title}{Joint subspace recovery and enhanced locality driven
  robust flexible discriminative dictionary learning}.
\newblock \bibinfo{journal}{IEEE Transactions on Circuits and Systems for Video
  Technology} .
\bibitem[{Zhang et~al.(2020)Zhang, Sun, Wang, Zha, Yan and
  Wang}]{zhang2020convolutional}
\bibinfo{author}{Zhang, Z.}, \bibinfo{author}{Sun, Y.}, \bibinfo{author}{Wang,
  Y.}, \bibinfo{author}{Zha, Z.}, \bibinfo{author}{Yan, S.},
  \bibinfo{author}{Wang, M.}, \bibinfo{year}{2020}.
\newblock \bibinfo{title}{Convolutional dictionary pair learning network for
  image representation learning}, in: \bibinfo{booktitle}{European Conference
  on Artificial Intelligence (ECAI)}.
\bibitem[{Zhang et~al.(2019c)Zhang, Sun, Zhang, Wang, Liu and
  Wang}]{zhang2019learning}
\bibinfo{author}{Zhang, Z.}, \bibinfo{author}{Sun, Y.}, \bibinfo{author}{Zhang,
  Z.}, \bibinfo{author}{Wang, Y.}, \bibinfo{author}{Liu, G.},
  \bibinfo{author}{Wang, M.}, \bibinfo{year}{2019}c.
\newblock \bibinfo{title}{Learning structured twin-incoherent twin-projective
  latent dictionary pairs for classification}, in: \bibinfo{booktitle}{IEEE
  International Conference on Data Mining (ICDM)}.
\bibitem[{Zou et~al.(2015)Zou, Ni, Zhang and Wang}]{zou2015deep}
\bibinfo{author}{Zou, Q.}, \bibinfo{author}{Ni, L.}, \bibinfo{author}{Zhang,
  T.}, \bibinfo{author}{Wang, Q.}, \bibinfo{year}{2015}.
\newblock \bibinfo{title}{Deep learning based feature selection for remote
  sensing scene classification}.
\newblock \bibinfo{journal}{IEEE Geoscience and Remote Sensing Letters}
  \bibinfo{volume}{12}, \bibinfo{pages}{2321--2325}.

\end{thebibliography}
%

%






\end{document}